\documentclass[10pt,twocolumn,letterpaper]{article}

\usepackage{cvpr}
\usepackage{multirow}
\usepackage{graphicx}
\usepackage{colortbl}
\usepackage{xcolor}
\usepackage{booktabs}
\usepackage{arydshln}
\usepackage{makecell}

\definecolor{myred}{rgb}{0.8, 0.0, 0.0} 
\definecolor{myblue}{rgb}{0.0, 0.4, 0.8} 
\definecolor{lightpeach}{rgb}{1, 0.9, 0.8}

\definecolor{cvprblue}{rgb}{0.21,0.49,0.74}
\usepackage[pagebackref,breaklinks,colorlinks,allcolors=cvprblue]{hyperref}
\usepackage[accsupp]{axessibility}

\def\paperID{3056} 
\def\confName{CVPR}
\def\confYear{2025}

\title{Vision-Language Gradient Descent-driven All-in-One Deep Unfolding Networks}

\hypersetup{
    colorlinks=true,
    urlcolor=magenta
}
\author{
Haijin Zeng$^{1,}$\thanks{Equal contribution. $\dagger$ Corresponding author.}\hspace{0.3cm}
Xiangming Wang$^{2, }$\footnotemark[1]\hspace{0.3cm}
Yongyong Chen$^{2,\dagger}$\hspace{0.2cm}
Jingyong Su$^{2,\dagger}$\hspace{0.2cm}
Jie Liu$^{2}$\\
$^{1}$ Harvard University\hspace{0.2cm} $^{2}$ Harbin Institute of Technology (Shenzhen) \\
Code: \url{github.com/xianggkl/VLU-Net}
}

\begin{document}
\maketitle
\begin{abstract}
Dynamic image degradations, including noise, blur and lighting inconsistencies, pose significant challenges in image restoration, often due to sensor limitations or adverse environmental conditions. 
Existing Deep Unfolding Networks (DUNs) offer stable restoration performance but require manual selection of degradation matrices for each degradation type, limiting their adaptability across diverse scenarios.
To address this issue, we propose the Vision-Language-guided Unfolding Network (VLU-Net), a unified DUN framework for handling multiple degradation types simultaneously.
VLU-Net leverages a Vision-Language Model (VLM) refined on degraded image-text pairs to align image features with degradation descriptions, selecting the appropriate transform for target degradation.
By integrating an automatic VLM-based gradient estimation strategy into the Proximal Gradient Descent (PGD) algorithm, VLU-Net effectively tackles complex multi-degradation restoration tasks while maintaining interpretability. 
Furthermore, we design a hierarchical feature unfolding structure to enhance VLU-Net framework, efficiently synthesizing degradation patterns across various levels.
VLU-Net is the first all-in-one DUN framework and outperforms current leading one-by-one and all-in-one end-to-end methods by 3.74 dB on the SOTS dehazing dataset and 1.70 dB on the Rain100L deraining dataset.
\end{abstract}    
\section{Introduction}
\label{sec:intro}

\begin{figure}[!t]
    \centering
    \includegraphics[width=1\linewidth]{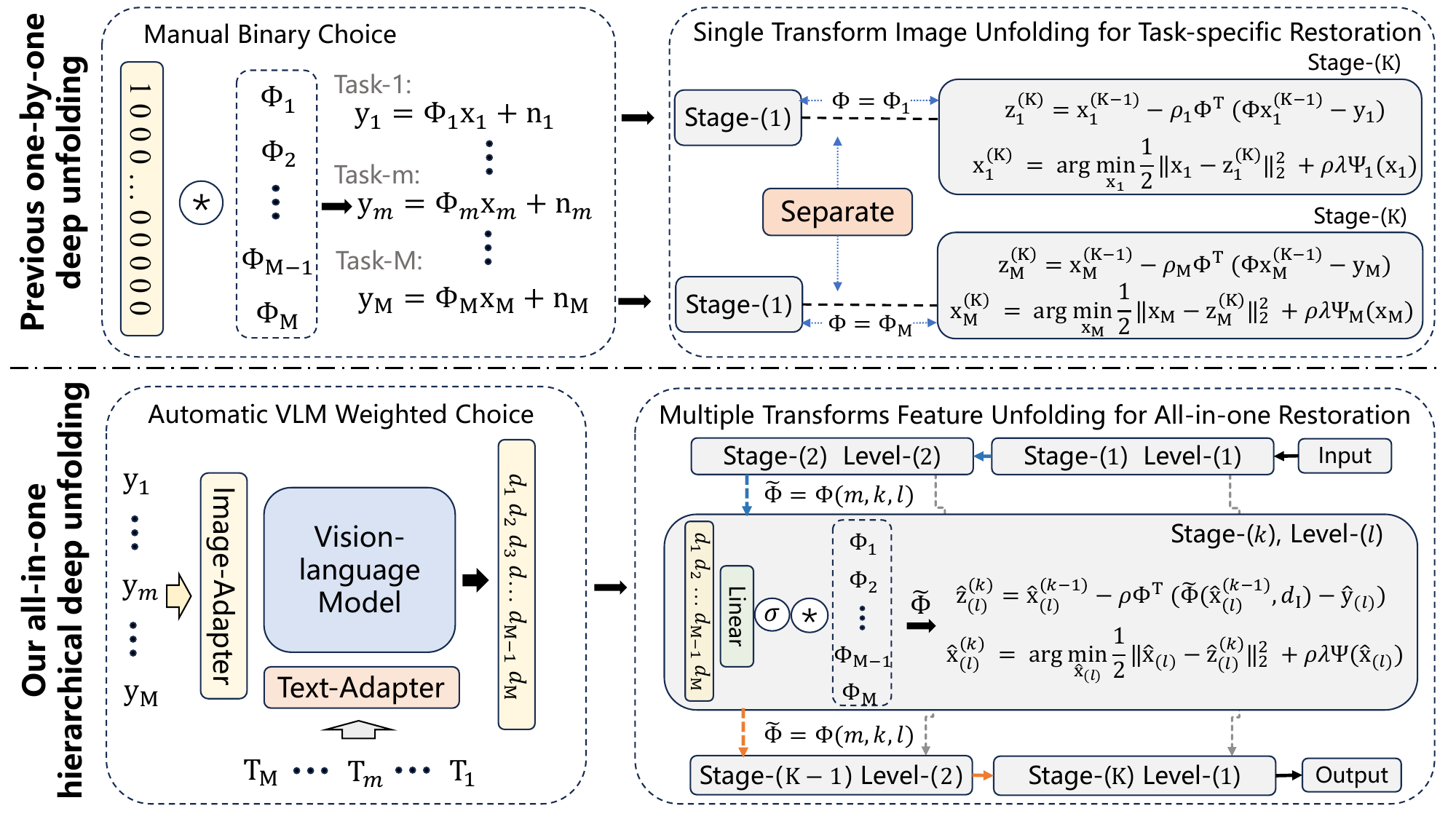}
    \vspace{-6mm}
    \caption{VLU-Net integrates an automatic gradient estimation strategy based on VLMs, enhancing existing DUNs which can only handle single task. This allows for simultaneous handling of multiple degradation without manual transform selection. Its hierarchical multi-stage unfolding structure efficiently synthesizes degradation patterns across various levels and stages.}
    \label{fig:highlight}
    \vspace{-4mm}
\end{figure}

Image restoration (IR) aims to recover original images $\mathbf{x}$, from degraded observations $\mathbf{y}$, affected by factors such as optical distortions \cite{zhang2013multi,chen2024hierarchical,hyun2017online,tsp} due to lens imperfections and shooting conditions, noise from photoelectric conversion, motion blur, and weather-induced artifacts like rain or haze. This can be formulated as an ill-posed equation:
\begin{equation} 
\label{eq:model}
    \mathbf{y} = \Phi \mathbf{x} + \mathbf{n},
\end{equation}
where $\mathbf{n}$ denotes the noise or errors in the imaging system and $\Phi$ represents the specific degradation matrix.
According to the degradation pattern defined by $\Phi$, numerous downstream tasks have been extensively studied, resulting in the evolution from model-based methods to deep learning-based approaches, and more recently to Deep Unfolding Networks (DUNs).
DUNs integrate interpretability and deep structural priors within a multi-stage framework, offering enhanced understanding and performance.

Model-based methods \cite{twist,gap_tv,NonLRMA,LLxRGTV,E3DTV} primarily rely on hand-crafted priors as regularizers, tailored for single task with specific type of degradation. 
While these methods are theoretically robust and interpretable, their reliance on precise prior information limits their flexibility in addressing multiple tasks with a range of degradation types. 
Additionally, since hand-crafted regularization terms like total variation for smoothness or nuclear norm for low-rankness are designed with specific priors in mind, model-based approaches often perform suboptimally when compared to deep learning methods \cite{mou2022deep,meng2023deep}, which leverage complex, data-driven priors and demonstrate superior adaptability.

Deep learning-based methods \cite{mst,cst,ddrm,dds2m,zamir2021multi}, utilizing informative data and efficient architectures, have successfully extracted richer and more generalized underlying patterns. 
They excel in single-degradation IR due to degradation-specific priors and targeted model designs. 
However, handling multiple degradation remains challenging, as interference among various degradation can compromise content extraction and degradation recognition.

To address this issue, several all-in-one restoration methods have emerged \cite{li2022all,zamir2022restormer,potlapalli2306promptir,zhang2023ingredient,cao2024hair,zhang2024perceive}. 
These approaches introduce unified frameworks for dealing with diverse degradation types by employing techniques such as prompt-based restoration \cite{potlapalli2306promptir}, contrastive loss for specialized modules \cite{li2022all}, and degradation-specific expert selection \cite{cao2024hair}.
Despite these innovations, deep learning-based models fundamentally operate as black boxes and lack interpretability, meaning the mechanisms driving their performance are not fully understood.
This lack of transparency limits insight into the operational dynamics of these models.

Building on model-based optimization approaches~\cite{bertero2021introduction}, DUNs translate the logic of iterative optimization algorithms, such as Proximal Gradient Descent (PGD)~\cite{beck2009fast} and Half-Quadratic Splitting (HQS)~\cite{wu2022uretinex, guo2023shadowdiffusion}, into a deep learning framework. 
DUNs achieve this by embedding deep modules as regularizers at each stage of the iterative process, facilitating the reconstruction of $\mathbf{x}$ from $\mathbf{y}$ with the degradation. 
This approach combines the interpretability of model optimization with the efficiency of deep structures, all within an end-to-end training paradigm.

Following the structure of PGD, IR problems in Eq. \eqref{eq:model} can be formulated into an iterative framework:
\begin{equation}
\label{gradient_descent}
    \mathbf{z}^{(k)} = \mathbf{x}^{(k-1)} - \rho\Phi^{\mathbf{T}}(\Phi \mathbf{x}^{(k-1)} - \mathbf{y}),
\end{equation}
\vspace{-4mm}
\begin{equation}
\label{proximal_mapping}
    \mathbf{x}^{(k)} = \arg\min_{\mathbf{x}} \frac{1}{2} \|\mathbf{x} - \mathbf{z}^{(k)}\|_2^2 + \rho\lambda \Psi(\mathbf{x}),
\end{equation}
where $k$ denotes the iteration step index, $\rho$ is the step size and $\Psi(\cdot)$ is the solver of image prior, which can be a handcrafted low-rank \cite{chang2020weighted} regularizer or in DUN architecture, a deep module \cite{mou2022deep} learned from extensive datasets. 
Eq. \eqref{gradient_descent}, the Gradient Descent Module (GDM), produces output from the prior iteration using the fixed degradation transform $\Phi$ defined in Eq. \eqref{eq:model}, integrating degradation information into the DUN model. Eq. \eqref{proximal_mapping}, the Proximal Mapping Module (PMM), updates the output $\mathbf{z}^{(k)}$ from the GDM and acts as a denoising step under Gaussian noise level $\sqrt{\rho\lambda}$ \cite{venkatakrishnan2013plug,ryu2019plug,zeng2020hyperspectral}. A typical DUN model comprises sequential stages, each containing one GDM and one PMM.

\emph{However, the specific degradation transform $\Phi$ in current DUNs, is basically a manual binary choice for different degradation matrices (as Figure \ref{fig:highlight}), preventing efficient handling of multiple degradations in an all-in-one IR approach.
Manually selecting the degradation matrix \cite{mou2022deep} leads to separate de-degradation processes for different tasks, requiring individual models for each, which lacks flexibility and poses challenges for lightweight deployment.}

To bridge this gap, a robust automatic mechanism for discerning multiple degradation types and levels is essential. 
Manual selection of specific degradation matrix in GDM is impractical for handling diverse scenarios, highlighting the need for a unified, adaptable approach.
Additionally, the sequential linkage of stages with fixed three-channel inputs causes the compression of essential content and degradation features, especially for the information estimation in GDM.
Furthermore, discrepancies in de-degradation and denoising levels for each GDM and PMM across stages, as shown in Eq. \eqref{gradient_descent} and \eqref{proximal_mapping}, emphasize the need for a hierarchical framework that integrates multi-level information, preserving and transmitting various content and degradation characteristics for effective restoration.


In this paper, we introduce a unified Vision-Language-guided Unfolding Network (VLU-Net) for comprehensive, all-in-one IR.
VLU-Net leverages high-dimensional degradation vectors derived from Vision-Language Models (VLMs), dynamically guiding both the generation and selection of degradation-specific information and relevant content features.
By harnessing the semantic and degradation understanding embedded within VLMs, VLU-Net can discern and adapt to multiple degradation types—such as noise, blur, and weather-induced distortions—in one instance without requiring separate models for each type. 
This framework effectively overcomes the limitations of traditional DUN methods, which often rely on pre-defined or general degradation matrices and struggle with generalization across scenarios with diverse degradation types and levels.
The primary contributions are outlined as follows:

\begin{itemize}
    \item We propose VLU-Net, the first all-in-one DUN for IR which uses VLM-derived high-dimensional features to automatically select degradation-aware keys, eliminating the need for pre-defined degradation knowledge.
    
    \item We design an efficient degradation fine-tuning strategy to improve feature alignment from encoders in high-dimensional vector space, leveraging the extraction capabilities of VLMs for precise degradation identification.
    
    \item VLU-Net's hierarchical structure unfolds optimization at feature level, integrating multi-level information to improve feature preservation and processing across stages.
    
    \item Experiments on single- and multi-degradation IR tasks demonstrate that VLU-Net achieves superior results over current state-of-the-art methods.
\end{itemize}
\section{Related Works}
\label{sec:related_works}
\noindent \textbf{Vision-Language Models.}
VLMs, such as those detailed in \cite{radford2021learning, jia2021scaling, li2022blip}, have demonstrated robust feature extraction and classification capabilities in high-dimensional spaces. These capabilities stem from their ability to produce generalizable visual and textual feature representations, which are crucial for a variety of downstream tasks. For instance, the Contrastive Language-Image Pre-training (CLIP) model introduced in \cite{radford2021learning} leverages contrastive learning to achieve remarkable alignment between images and their corresponding textual descriptions. Building upon this, BLIP \cite{li2022blip} enhances pretraining effectiveness through the generation of synthetic captions. 
Additionally, DA-CLIP \cite{luo2024controlling, luo2024photo} incorporates degraded image-text pairs and integrates these with diffusion models to facilitate image restoration. 
Inspired by these pioneering studies, we employ CLIP (tuned with image-text degradation pairs) to represent visual degradation features, which in turn guide the training process of our comprehensive all-in-one DUN model.

\noindent \textbf{Deep Unfolding Networks for Image Restoration.}
Under the framework of logical optimization and deep learning architectures, DUN enhances IR by unfolding iterative functions into an end-to-end neural network. 
This approach incorporates degradation-specific transforms and multi-stage noise removal capabilities. 
For example, ISTA-Net \cite{tra_3} employs a sparse nonlinear architecture based on the iterative shrinkage thresholding algorithm for image Compressive Sensing (CS). 
Similarly, ADMM-Net \cite{admm-net} introduces the Alternating Direction Method of Multipliers (ADMM) algorithm, using a deep tensor-singular value decomposition structure to address video snapshot compressive imaging. 
DGUNet \cite{mou2022deep} further builds on these concepts by integrating a flexible GDM for dynamic, manual selection of the degradation matrix, applicable to image deblurring and CS. 
Additionally, UFC-net \cite{wang2024ufc} enhances feature extraction between stages with fixed-point optimization. 
\emph{Despite these advancements, existing DUN methods remain largely degradation-specific, requiring separate models for each restoration task.} 
Furthermore, the sequential linkage with image level compression often limits the extraction of multi-level information, hindering their capacity for comprehensive feature representation.

\noindent \textbf{All-in-one Image Restoration.}
All-in-one IR seeks to develop a unified framework that addresses various degradations in a single image processing workflow.
Recent advancements have introduced multiple methodologies that enhance task-specific information extraction, each contributing uniquely to this unified restoration objective.
AirNet \cite{li2022all} uses contrastive learning to differentiate inputs from various tasks, guiding appropriate processing paths.
PromptIR \cite{potlapalli2306promptir} employs a prompt generation module that extracts task-specific information for each input, integrating these prompts into the main restoration model. 
NDR \cite{yao2024neural} introduces a degradation query module that utilizes attention for dynamic degradation representation under a bidirectional optimization strategy.
InstructIR \cite{conde2024high} leverages natural language prompts to guide the restoration of degraded images with human-written instructions.
While these methods employ innovative strategies to identify degradation types and levels, they primarily focus on multi-modal information extraction, lacking interpretable architectures such as DUN for comprehensive degradation analysis.
\section{Methods}

\begin{figure*}[!t]
    \centering
    \includegraphics[width=1\linewidth]{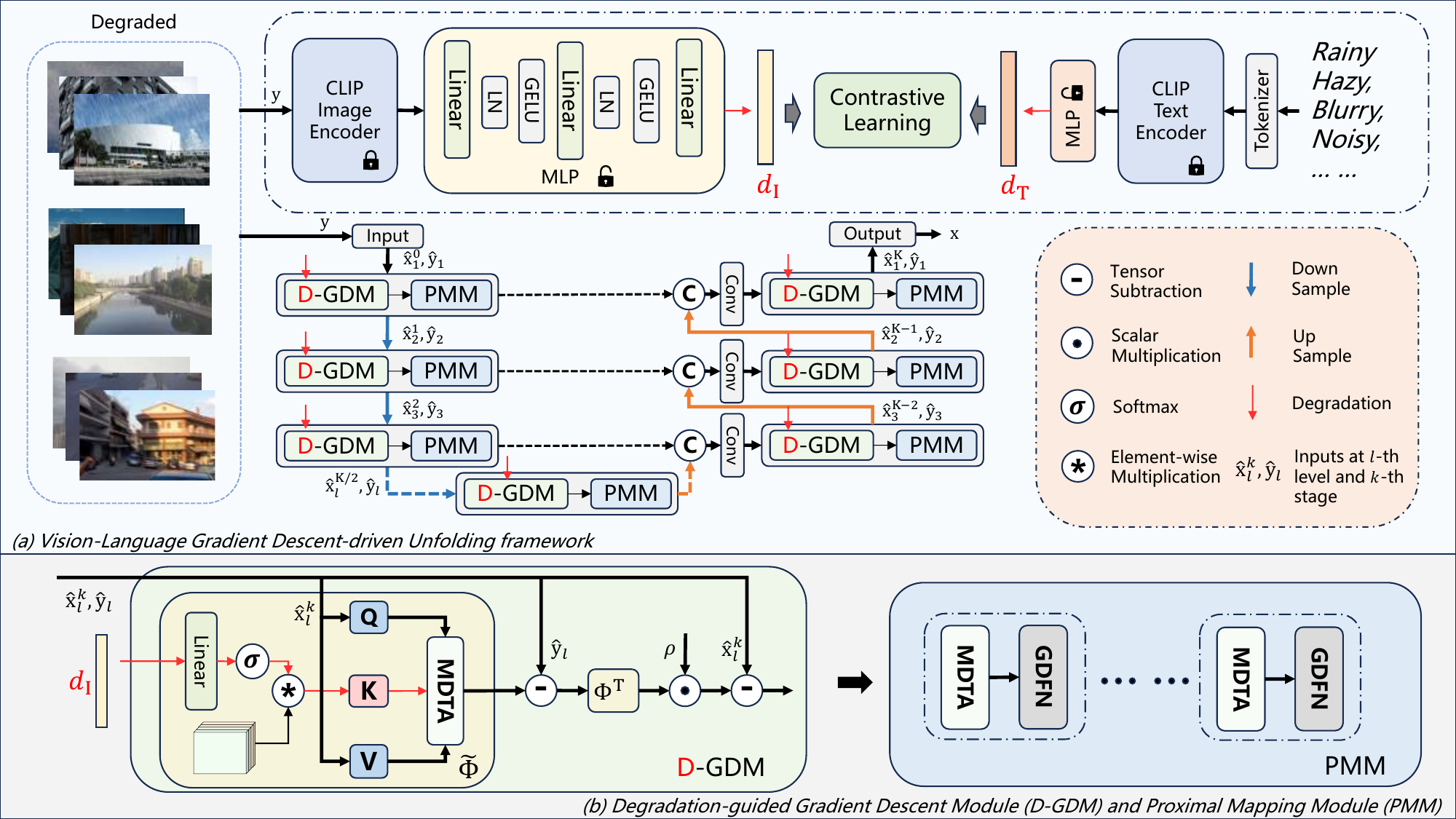}
    \vspace{-5mm}
    \caption{Overview of our VLU-Net, an all-in-one hierarchical DUN for multiple degradations, including fine-tuning of CLIP and the primary IR process. Fine-tuning phase employs contrastive learning with degradation image-text pairs, while the IR process involves a $\mathbf{K}$-stage DUN with $l$ levels, projection and back projection. Each stage utilizes a Degradation-guided GDM (D-GDM) and PMM to adaptively handle degraded inputs and maintain multi-level information across stages. Noted detailed PMM is presented in supplementary materials.}
    \label{fig:overview}
    \vspace{-4mm}
\end{figure*}

\subsection{Motivation and Overview}
\label{sec:motivation}
\noindent \textbf{Motivation.}
Derived from iterative optimization algorithms, Eq. \eqref{gradient_descent} and \eqref{proximal_mapping} describe two operations in the $k$-th stage of DUN.
GDM refines the stage image $\mathbf{x}^{(k)} \in \mathbb{R}^{H \times W \times 3}$ based on the previous stage output $\mathbf{x}^{(k-1)}$ and original degraded image $\mathbf{y} \in \mathbb{R}^{H \times W \times 3}$ through the degradation equation with transform $\Phi(\cdot)$. 
Then PMM maps the GDM output into next stage by solving a subproblem that is equivalent to a denoising task with Gaussian noise level $\sqrt{\rho\lambda}$. 
By addressing the following two issues, we aim to fully leverage GDM's potential for efficient all-in-one IR.

Firstly, the degradation transform in current DUNs is inflexible and requires manual selection. 
Specifically, existing DUNs either rely on a fixed degradation transform or introduce a learnable parametric transform to approximate unknown degradations as in \cite{tra_3, mou2022deep}, which allows a single model to be trained multiple times for multiple tasks.
However, determining a specific transform for each task in advance is impractical for real-world images, where degradation is often unknown. 
On the other hand, with learnable transforms, degradation information may be overlooked if no prior constraints are enforced, leading to a mixed degradation transform that cannot accurately address individual tasks. 
Therefore, an automatic degradation detector and classifier is required to effectively guide the GDM, enabling a unified all-in-one IR framework (where a single model is trained once to handle multiple tasks).


We leverage a multi-modal vision-language model to learn degradation information, enabling an all-in-one DUN. Specifically, we customize a pre-trained degradation-aware multi-modal VLM and fine-tune it using degraded images paired with corresponding degradation descriptions (See Sec.~\ref{sec:VLGD} for details). 
The accuracy of degradation learning is supervised by measuring the similarity (see Sec.~\ref{sec:abalation} for results) between the model's outputs and the corresponding textual labels, such as ``image with noise" or ``image with rain".
Given the tested degraded image $\mathbf{y}$ and text labels $\mathbf{T} \in \mathbb{R}^{\mathbf{M} \times L}$ where $\mathbf{M}$ is the number of the degradations and $L$ is the length of the text embedding, we have:
\begin{equation}
    s_i = \frac{e^{\gamma \cdot \operatorname{cos}(\operatorname{E_I}(\mathbf{y}), \operatorname{E_T}(\mathbf{T}_i))}}
{\sum_{j=1}^{\mathbf{M}} e^{\gamma \cdot \operatorname{cos}(\operatorname{E_I}(\mathbf{y}), \operatorname{E_T}(\mathbf{T}_j))}},
\end{equation}
where $\operatorname{E_I}(\cdot)$ and $\operatorname{E_T}(\cdot)$ are the image encoder and text encoder of CLIP, $\operatorname{cos}(\cdot, \cdot)$ is the cosine similarity calculation and $\gamma$ is the scaling factor to sharpen the probabilities.

CLIP can classify the degradation except for different noise levels and after fine-tuning CLIP shows a stronger representation ability for each degradation.
Thus with the degradation-aware embedding, we can construct an automatic modulation for different types of degradation transforms, offering a more flexible and generalized GDM.

Secondly, existing DUNs treat inputs homogeneously across stages, especially within the GDM.
At each stage, only a three-channel stage image $\mathbf{x}^{(k-1)}$ and original degraded image $\mathbf{y}$ are processed, while within the PMM, high-dimensional features are briefly introduced but subsequently compressed for the next stage. 
Thus this repeated compression-decompression constrains the model’s ability to capture nuanced degradation features and intricate image structures, particularly in deeper stages, creating a bottleneck that hinders DUN scalability.
Moreover, each GDM processes similar low-dimensional information, though different stages should ideally address distinct situations, such as varying denoising levels as indicated by Eq. \eqref{proximal_mapping}. 
To bridge this gap, we embed $\mathbf{y}$ into a high-dimensional feature space with a linear transformation (also the inverse), constructing a hierarchical DUN that promotes multi-level features across stages for enhanced information handling.

\noindent \textbf{Overview.}
We present the main framework in Figure \ref{fig:overview}, which comprises two main procedures: the fine-tuning process for CLIP and the primary IR process with our VLU-Net.
For the fine-tuning phase, we compile $\mathbf{M}$ degradation datasets to enhance CLIP’s capacity for representing various degradations. 
Each batch includes $B$ degraded images $\mathbf{Y} \in \mathbb{R}^{B \times H \times W \times 3}$ and a corresponding list of degradation descriptions $\hat{\mathbf{T}} \in \mathbb{R}^{B \times L}$.
CLIP is fine-tuned using a contrastive learning strategy, with two additional Multi-Layer Perceptrons (MLPs) serving as learnable layers.
In the IR process, VLU-Net begins by applying a linear transformation to project the input data to a feature level. The transformed inputs are then processed through a $\mathbf{K}$-stage DUN and subsequently projected back to the image level using another 
ideally invertible linear transformation.
Each $k$-th stage within the $l$-th level consists of a Degradation-guided GDM (D-GDM) and a PMM which has several Transformer blocks, Multi-Dconv head Transposed Attention (MDTA) and Gated-Dconv Feed-forward Network (GDFN).

Stage features and degraded inputs are propagated through hierarchical stages using down-sampling or up-sampling operations, ensuring that multi-level degraded and content information is preserved and expanded across different stages.
Each D-GDM requires a stage input $\hat{\mathbf{x}}^{(k-1)}_{(l)} \in \mathbb{R}^{H_{(l)} \times W_{(l)} \times C_{(l)}}$, a degraded input $\hat{\mathbf{y}}_{(l)} \in \mathbb{R}^{H_{(l)} \times W_{(l)} \times C_{(l)}}$—where $H_{(l)}$, $W_{(l)}$, and $C_{(l)}$ represent the height, width, and channels at the $l$-th level—and a $D$-dimensional degradation-aware vector $d_{\mathbf{I}} \in \mathbb{R}^{D}$ obtained from the CLIP image encoder.
As formulated in Eq. \eqref{gradient_descent}, the D-GDM applies the proposed VLM-guided cross-attention module $\tilde{\Phi}$ as dynamic, degradation-aware transform, and a regular attention module as $\Phi^{\mathbf{T}}$ at each stage.
The degradation vector $d_{\mathbf{I}}$, processed through the frozen fine-tuned CLIP model, is projected and softmaxed to serve as an automatic VLM weighted retrieval vector, selecting corresponding learnable degradation key effectively.

\subsection{Vision-Language-driven Gradient Descent} \label{sec:VLGD}
Below we describe the proposed D-GDM about how we leverage CLIP for the dynamic detection of various degradation and how this VLM be induced into GDM.
The fine-tuning process of CLIP is shown in the top of Figure \ref{fig:overview} (a).
The structure of the D-GDM is shown in Figure \ref{fig:overview} (b).

Based on the pre-trained CLIP which trained with a large set of image-text content pairs, We adopt $\mathbf{M}$ degraded datasets and corresponding degraded texts to enhance its degradation recognition.
We add two learnable modules and each module is a three-layer MLP with layer normalization and Gaussian Error Linear Unit (GELU) \cite{hendrycks2016gelu}, in front of the image and text encoders of CLIP as adapters.
Concretely, we computes the contrastive learning loss as:
\begin{equation}
\label{loss_single_sample}
    \mathcal{L}(\mathbf{Y}, \mathbf{\hat{T}}) = -\frac{1}{B} \sum_{i=1}^{B} \log \left( \frac{ e^{\tau \operatorname{cos}(\hat{\operatorname{E_I}}(\mathbf{Y}_i),\hat{\operatorname{E_T}}(\mathbf{\hat{T}}_i))}}{\sum_{j=1}^{B} e^{ \tau \operatorname{cos}(\hat{\operatorname{E_I}}(\mathbf{Y}_i),\hat{\operatorname{E_T}}(\mathbf{\hat{T}}_j))}} \right),
\end{equation}
where $\tau$ is the learnable temperature parameter which controls the contrastive strength, $\hat{\operatorname{E_I}}(\cdot)$ and $\hat{\operatorname{E_T}}(\cdot)$ represent the CLIP image encoder and text decoder with MLPs.
Thus we obtain a degradation-aware VLM with high-dimensional output for degradation analysis, and:
\vspace{-1mm}
\begin{equation}
    d_{\mathbf{I}} = \hat{\operatorname{E_I}}(\mathbf{y}),
\vspace{-1mm}
\end{equation}
where $\mathbf{y}$ is the original degraded image.

To induce the degradation prior into the GDM, we propose the $\tilde{\Phi}$, a degradation-attention module which dynamically selects the learnable keys for different degradation types and levels.
This module projects the degradation vector $d_{\mathbf{I}}$ from $\hat{\operatorname{E_I}}(\cdot)$ to low-dimensional space and passes through a softmax operation $\sigma(\cdot)$ to obtain a degradation retrieval vector.
With channel-wise multiplication, this vector extracts corresponding degradation keys from a learnable degradation key database $K_{(l)}$ at $l$-th level (See Sec.~\ref{sec:HDUN} for ``$l$-th level''), which contains several degradation basis tensors.
This strategy can not only leverage the VLM for automatic selection of the corresponding degradation transform, but also integrate degradation information from other degradation modes, allowing for joint judgment and restoration for target data instance.
Then this VLM weighted degradation key and degraded inputs are computed by a multi-head attention module.
Relevant equations can be formulated as:
\vspace{-3mm}
\begin{equation}
\label{key}
    \operatorname{key} = \operatorname{sum}(\sigma(\operatorname{Linear}(d_{\mathbf{I}})) * K_{(l)}),
\end{equation}
\vspace{-4mm}
\begin{equation}
    \label{D-GDM}
    \tilde{\Phi}(\hat{\mathbf{x}}^{(k-1)}_{(l)}, d_{\mathbf{I}}) = \operatorname{MDTA}(\hat{\mathbf{x}}^{(k-1)}_{(l)}, \operatorname{key}, \hat{\mathbf{x}}^{(k-1)}_{(l)}).
\end{equation}
Following GDM in Eq. \eqref{gradient_descent}, we transform back the residuals between degradation-aware features and degraded embedding by $\Phi^{\mathbf{T}}$, which is a regular self-attention module for feature refinement.
Finally, D-GDM adjust the stage inputs with the refined residuals and a strength controller $\rho$.

\subsection{Hierarchical DUN Architecture}
\label{sec:HDUN}
Below we present our hierarchical DUN about how we process the stage image $\mathbf{x}^{(k-1)}$ in feature level and how we deal with the degraded input $\mathbf{y}$ under different levels.
Figure \ref{fig:overview} shows the main framework of our VLU-Net.

To explore the degradation processing abilities of GDM, we aim to embed the degraded images into feature level with a linear transform $\mathbf{W} \in \mathbb{R}^{3 \times C_{(1)}}$ and after the last stage, project them back with its inverse $\mathbf{W}^{-1} \in \mathbb{R}^{C_{(1)} \times 3}$ ideally, achieving a level-transform strategy.
We utilize two linear convolution operations to learn the transforms and:
\vspace{-2mm}
\begin{equation}
    \label{lineartrans_model}
    \mathbf{y} \times_3 \mathbf{W} = (\Phi \mathbf{x} + \mathbf{n}) \times_3 \mathbf{W},
\end{equation}
\vspace{-6mm}
\begin{equation}
    \label{lineartransFinal_model}
    \hat{\mathbf{y}} = \Phi \hat{\mathbf{x}} + \hat{\mathbf{n}},
\end{equation}
\vspace{-6mm}
where $\times_3$ denotes the mode-3 product \cite{kolda2009tensor} and
\vspace{4mm}
\begin{equation}
    \label{lineartrans_model_back}
    \mathbf{x} = \hat{\mathbf{x}} \times_3 \mathbf{W}^{-1}.
\vspace{-2mm}
\end{equation}
Thus optimization can be unfolded in feature level, removing restriction of dimensional compression between stages and expanding high-dimensional features into D-GDM.

The aforementioned transforms also serve as the cornerstone for applying various degraded inputs at different levels,
\eg, $\hat{\mathbf{y}}_{(l)}$ at $l$-th level.
Current DUNs need the same original degraded three-channel image $\mathbf{y}$ for each stage, and if we directly sample or generate the input in image level, we directly loss the actual pixel information.
Under the feature level embedding, we down sample and up sample the degraded input $\mathbf{y}$ to different feature levels from a high-dimensional space, leveraging more details and especially the multi-level degraded information for GDM.
Additionally, we also add the residual connections for the degraded inputs in the same level, which preserves and transfers original degraded features for later stages.
Thus we make each stage process hierarchical information from different levels in DUN framework.
And for the $k$-th stage within the $l$-th level, our hierarchical degradation unfolding procedure is:
\vspace{-2mm}
\begin{equation}
\label{gradient_descent_lineartrans}
    \hat{\mathbf{z}}^{(k)}_{(l)} = \hat{\mathbf{x}}^{(k-1)}_{(l)} - \rho \Phi^{\mathbf{T}}(\tilde{\Phi} (\hat{\mathbf{x}}^{(k-1)}_{(l)}, d_{\mathbf{I}}) - \hat{\mathbf{y}}_{(l)}),
\end{equation}
\vspace{-7mm}
\begin{equation}
\label{proximal_mapping_lineartrans}
    \hat{\mathbf{x}}^{(k)}_{(l)} = \arg\min_{\hat{\mathbf{x}}_{(l)}} \frac{1}{2} \| \hat{\mathbf{x}}_{(l)} - \hat{\mathbf{z}}^{(k)}_{(l)} \|_2^2 + \rho\lambda \Psi(\hat{\mathbf{x}}_{(l)}).
\end{equation}
\section{Experiments}

\begin{table*}
\centering
\def\arraystretch{1.1}
\setlength{\tabcolsep}{4pt}
\caption{Comparison of dehazing, deraining, denoising, deblurring and low-light enhancement results under NHRBL settings. The PSNR values (highlighted in orange cell) and SSIM scores are reported, with the highest value marked in \textcolor{myred}{Red} and the second-highest in \textcolor{myblue}{Blue}.}
\vspace{-2mm}
\scalebox{0.8}{
\begin{tabular}{llll>{\columncolor{lightpeach!50}}cc>{\columncolor{lightpeach!50}}cc>{\columncolor{lightpeach!50}}cc>{\columncolor{lightpeach!50}}cc>{\columncolor{lightpeach!50}}cc>{\columncolor{lightpeach!50}}cc>{\columncolor{lightpeach!50}}c}
\toprule
\multirow{2}{*}{\text{Type}} & \multirow{2}{*}{\text{Method}} & \multirow{2}{*}{Reference} & \multirow{2}{*}{\text{Params.}} & \multicolumn{2}{c}{\text{Dehazing}} & \multicolumn{2}{c}{\text{Deraining}} &\multicolumn{2}{c}{\text{Denoising}} & \multicolumn{2}{c}{\text{Delurring}} & \multicolumn{2}{c}{\text{Low-light}} & \multicolumn{2}{c}{\text{Average}} \\
\cmidrule(l{0.7em}r{0.7em}){5-6}
\cmidrule(l{0.7em}r{0.7em}){7-8}
\cmidrule(l{0.7em}r{0.7em}){9-10}
\cmidrule(l{0.7em}r{0.7em}){11-12}
\cmidrule(l{0.7em}r{0.7em}){13-14}
\cmidrule(l{0.7em}r{0.7em}){15-16}
& & & & \multicolumn{2}{c}{\textit{SOTS}} & \multicolumn{2}{c}{\textit{Rain100L}} & \multicolumn{2}{c}{\textit{CBSD68}$_{\sigma=25}$} & \multicolumn{2}{c}{\textit{GoPro}} & \multicolumn{2}{c}{\textit{LOL}} & \multicolumn{1}{c}{\text{PSNR}} & \multicolumn{1}{c}{\text{SSIM}}\\
\toprule
\multirow{3}{*}{\makecell{ One-by-one}} 
 & DGUNet \cite{mou2022deep} & CVPR’22 & 17M &24.78&.940 &36.62&.971 &31.10&.883&27.25&.837& 21.87&.823& 28.32&.891\\
& Restormer \cite{zamir2022restormer} &  CVPR’22 & 26M &24.09&.927 &34.81&.962 &31.49&.884&27.22&.829 &20.41&.806 &27.60&.881\\
 & MambaIR \cite{guo2024mambair} & CVPR’24 & 27M & 25.81&.944 &36.55&.971 & 31.41&.884&28.61&.875 &22.49&.832 &28.97&.901\\
\toprule
\multirow{8}{*}{\makecell{All-in-one \\ (end-to-end)}} & AirNet \cite{li2022all} & CVPR’22& 9M &21.04&.884 &32.98&.951& 30.91&.882& 24.35&.781 &18.18&.735 &25.49&.846 \\
 & Transweather \cite{valanarasu2022transweather} & CVPR’22 & 38M &21.32&.885& 29.43&.905&29.00&.841 & 25.12&.757 &21.21&.792& 25.22&.836 \\
 & IDR \cite{zhang2023ingredient} & CVPR’23  & 15M &25.24&.943& 35.63&.965&\textcolor{myred}{\textbf{31.60}}&\textcolor{myblue}{\textbf{.887}}& 27.87&.846 &21.34&.826 &28.34&.893 \\
 & PromptIR \cite{potlapalli2306promptir} & NeurIPS’23 & 33M &26.54&.949 &36.37&.970 & \textcolor{myblue}{\textbf{31.47}}&.886&28.71&.881 &\textcolor{myblue}{\textbf{22.68}}&.832& 29.15&.904 \\
 & Gridformer \cite{wang2024gridformer} & IJCV’24 & 34M & 26.79&.951 &36.61&.971&31.45&.885 &\textcolor{myblue}{\textbf{29.22}}&\textcolor{myblue}{\textbf{.884}} &22.59&.831& 29.33&.904\\
 &  InstructIR \cite{conde2024high} & ECCV’24 & 16M&\textcolor{myblue}{\textbf{27.10}}&\textcolor{myblue}{\textbf{.956}}&\textcolor{myblue}{\textbf{36.84}}&\textcolor{myblue}{\textbf{.973}}&31.40&\textcolor{myblue}{\textbf{.887}}&\textcolor{myred}{\textbf{29.40}}&\textcolor{myred}{\textbf{.886}}&\textcolor{myred}{\textbf{23.00}}&\textcolor{myred}{\textbf{.836}}&\textcolor{myblue}{\textbf{29.55}}&\textcolor{myred}{\textbf{.907}} \\
 \cmidrule(l{0.7em}r{0.7em}){2-16}
\rowcolor{lightgray!20}
\cellcolor{white} (deep unfolding) & VLU-Net & Ours & 35M &  \textcolor{myred}{\textbf{30.84}} &  \textcolor{myred}{\textbf{.980}} &  \textcolor{myred}{\textbf{38.54}} &  \textcolor{myred}{\textbf{.982}} & 31.43 &  \textcolor{myred}{\textbf{.891}}& 27.46 & .840& 22.29&\textcolor{myblue}{\textbf{.833}}& \textcolor{myred}{\textbf{30.11}}& \textcolor{myblue}{\textbf{.905}}\\

    
\bottomrule
\end{tabular}
}
\label{NHRBL}
\vspace{-3mm}
\end{table*}

\subsection{Experimental Setup}
\label{experiments}

\noindent \textbf{Datasets.}
Following \cite{li2022all,potlapalli2306promptir,zhang2024perceive}, we prepare the datasets for one-by-one and all-in-one IR.
For one-by-one IR settings, we combine the \textit{BSD400} and \textit{WED} datasets (400 and 4744 training images), adding Gaussian noise at level $\sigma \in \{ 15,25,50 \}$ for image denoising, and testing at \textit{CBSD68} and \textit{Urban100} dataset.
For image dehazing, we adopt the \textit{RESIDE-$\beta$-OTS} dataset which has 72135 images for training and the \textit{SOTS-Outdoor} dataset with 500 samples for testing.
For image deraining, we use the \textit{Rain100L} dataset and it contains 200 clean-rainy image pairs for training and 100 pairs for testing.
For image deblurring, \textit{GoPro} dataset is utilized which consists of 2103 training images and 1111 testing images.
For low-light enhancement, we leverage the \textit{LOL} dataset with 485 images for training and 15 images for testing.
Under one-by-one IR, we train and test using datasets from single degradation at a time.
For all-in-one IR settings, we list two common mixed degradation: ``Noise+Haze+Rain" as NHR and ``Noise+Haze+Rain+Blur+Low-light" as NHRBL.
We train one model on mixed datasets with all corresponding training images (cropped to $128 \times 128$ and randomly flipped horizontally and vertically) from multiple degradation and test one-by-one on single degradation dataset.

\noindent \textbf{Metrics.}
For evaluation, we present the Peak Signal to Noise Ratio (PSNR) and the Structural Similarity (SSIM).

\noindent \textbf{Baselines.}
For one-by-one IR methods, we compared our VLU-Net with eight specific IR (trained single time for single task) methods (DnCNN \cite{zhang2017beyond}, FFDNet \cite{zhang2018ffdnet}, DehazeNet \cite{cai2016dehazenet}, EPDN \cite{qu2019enhanced}, FDGAN \cite{dong2020fd}, UMR \cite{yasarla2019uncertainty}, SIRR \cite{wei2019semi}, MSPFN \cite{jiang2020multi}) and six general IR (trained multiple times for multiple tasks) methods (MPRNet \cite{zamir2021multi}, SwinIR \cite{liang2021swinir}, DGUNet \cite{mou2022deep}, FSNet \cite{cui2023image}, Restormer \cite{zamir2022restormer}, MambaIR \cite{guo2024mambair}).
And seven all-in-one IR (trained single time for multiple tasks) methods (AirNet \cite{li2022all}, Transweather \cite{valanarasu2022transweather}, IDR \cite{zhang2023ingredient}, PromptIR \cite{potlapalli2306promptir}, NDR \cite{yao2024neural}, Gridformer \cite{wang2024gridformer}, InstructIR \cite{conde2024high}) are adopted.
Especially, our method is the only all-in-one method within DUN architecture.

\noindent \textbf{Implementation Details.}
All experiments are conducted on 8 NVIDIA GeForce RTX 4090 GPUs.
The pre-trained CLIP with ViT-B/32 acts as backbone and be fine-tuned for 100 epochs with AdamW optimizer ($1e-5$ learning rate and 32 batch size).
The frozen image encoder of CLIP requires 88M Params. and 18GMACs.
Our VLU-Net, optimized with $L_1$ loss, consists eight non-shared stages for four levels and the number of transformer blocks \{4,6,6,8\} for PMM in each stage.
Details of the transformer block we used are provided in supplementary materials.
We leverage the AdamW optimizer with $\beta_1$=0.9 and $\beta_2$=0.999, and the learning rate is $2e-4$ following a 15-epoch linear warm up and cosine annealing strategy.
For one-by-one IR, the batch size is 8 and for all-in-one IR, it tunes to 32 for 200 epochs.

\vspace{-1mm}
\subsection{Experiment Results}
\vspace{-1mm}

\begin{figure*}
    \centering
    \setlength{\tabcolsep}{1pt}
    \scalebox{1.0}{
    \begin{tabular}{lccccc}
    \multirow{2}{*}{\rotatebox{90}{Dehazing}}&
        \includegraphics[width=0.18\linewidth]{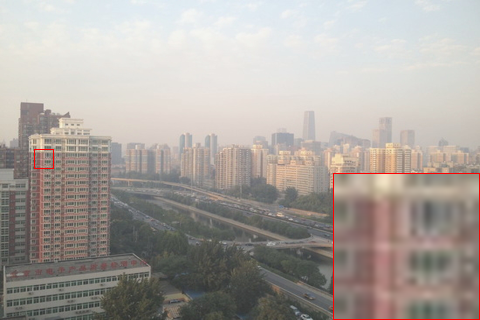} &
        \includegraphics[width=0.18\linewidth]{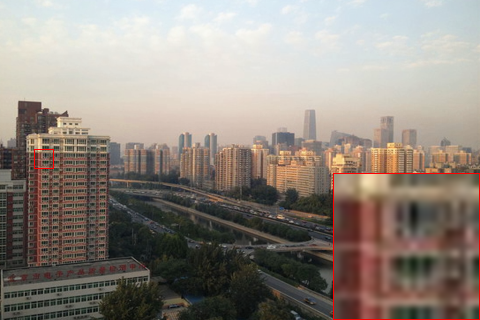} &
        \includegraphics[width=0.18\linewidth]{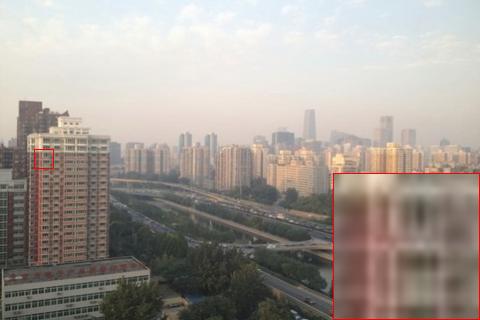} &
        \includegraphics[width=0.18\linewidth]{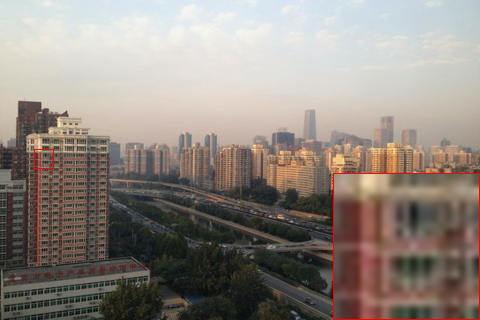} &
        \includegraphics[width=0.18\linewidth]{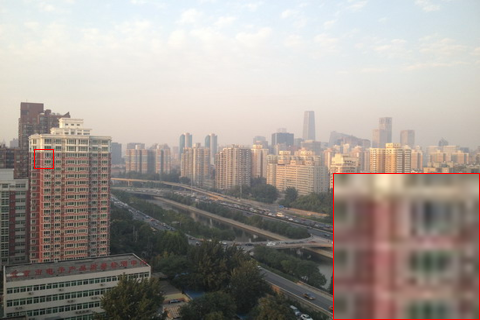}\\
        &\includegraphics[width=0.18\linewidth]{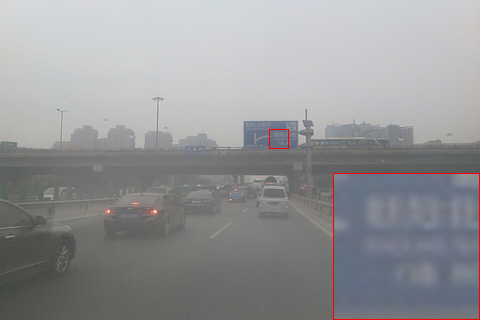} &
        \includegraphics[width=0.18\linewidth]{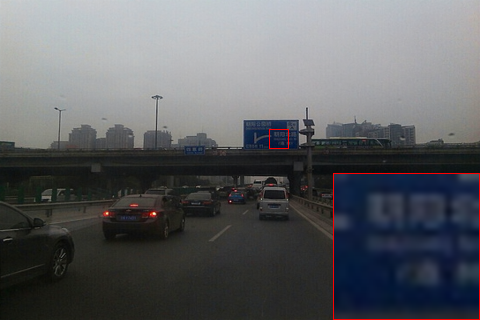} &
        \includegraphics[width=0.18\linewidth]{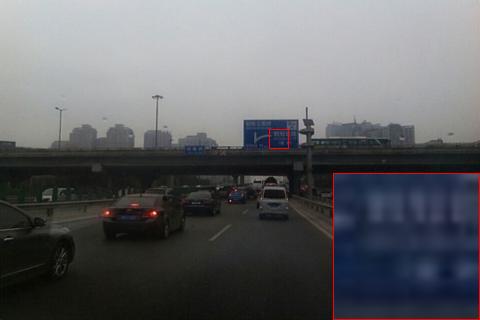} &
        \includegraphics[width=0.18\linewidth]{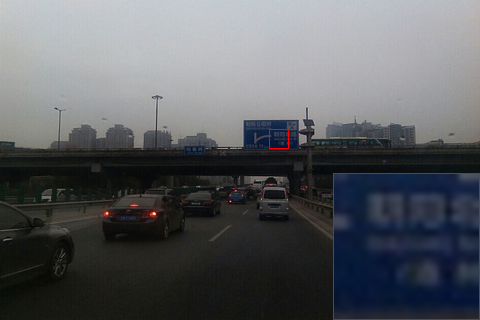} &
        \includegraphics[width=0.18\linewidth]{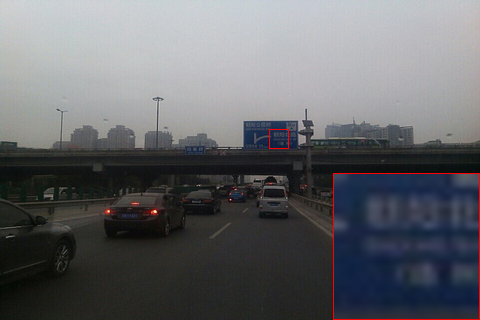}\\
        \hdashline[2pt/1pt]
        \noalign{\vskip 3.5pt}
        \multirow{2}{*}{\rotatebox{90}{Deraining}}&
        \includegraphics[width=0.18\linewidth]{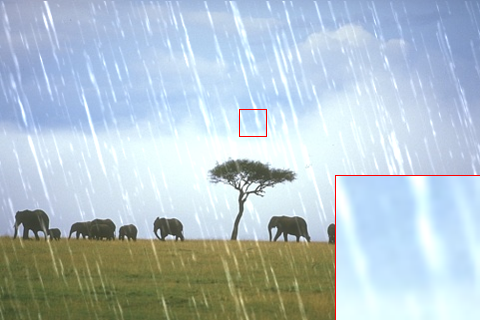} &
        \includegraphics[width=0.18\linewidth]{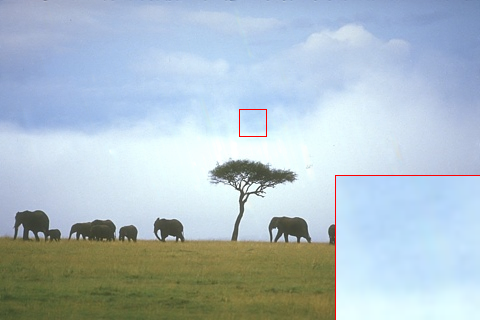} &
        \includegraphics[width=0.18\linewidth]{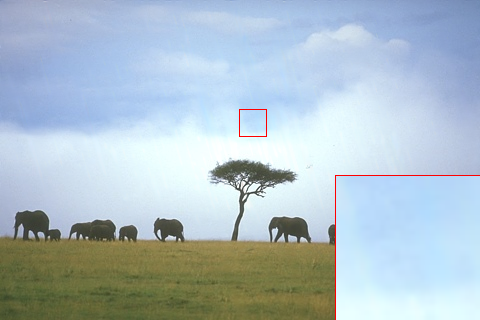} &
        \includegraphics[width=0.18\linewidth]{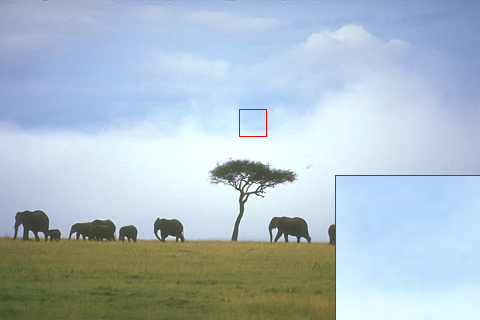} &
        \includegraphics[width=0.18\linewidth]{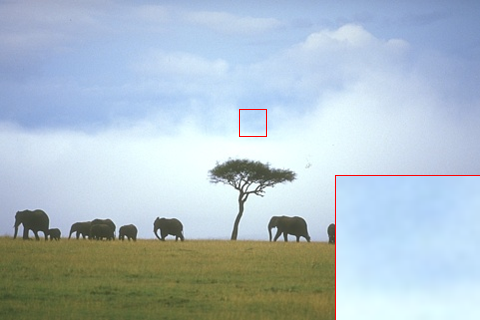}\\
        &\includegraphics[width=0.18\linewidth]{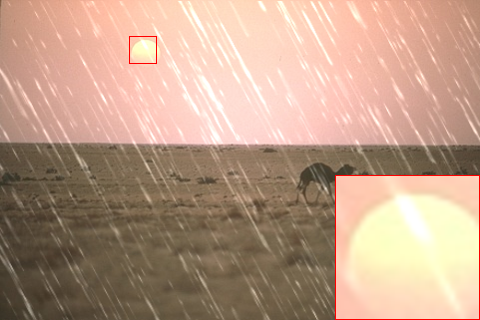} &
        \includegraphics[width=0.18\linewidth]{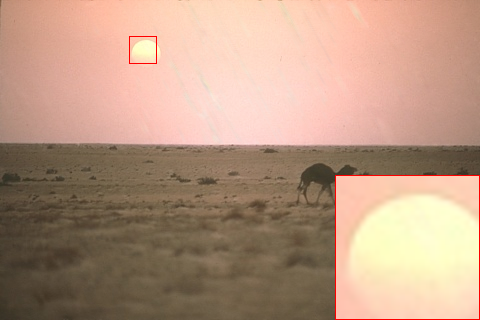} &
        \includegraphics[width=0.18\linewidth]{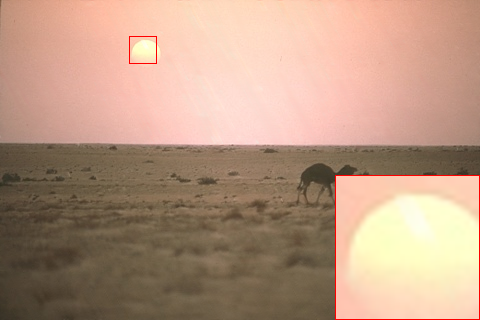} &
        \includegraphics[width=0.18\linewidth]{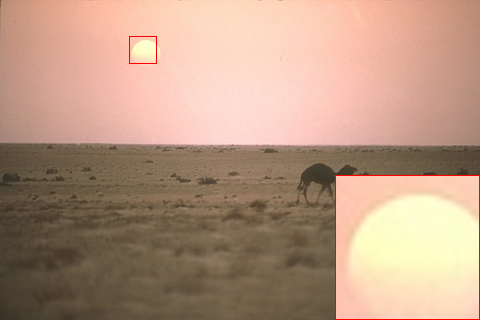} &
        \includegraphics[width=0.18\linewidth]{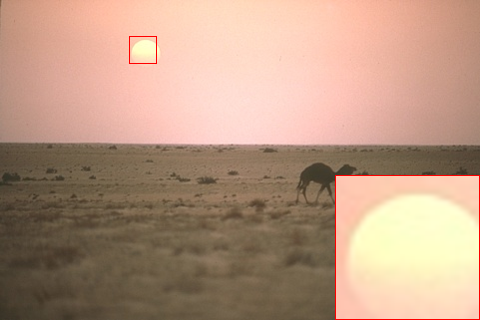}\\
        \hdashline[2pt/1pt]
        \noalign{\vskip 3.5pt}
        \multirow{2}{*}{\rotatebox{90}{Denoising}}&
        \includegraphics[width=0.18\linewidth]{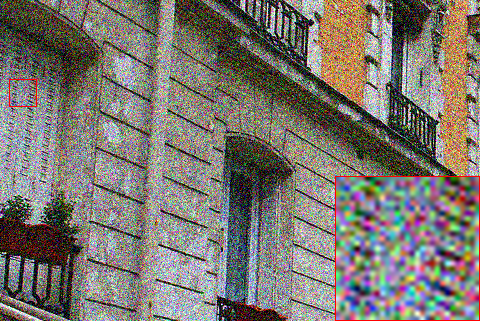} &
        \includegraphics[width=0.18\linewidth]{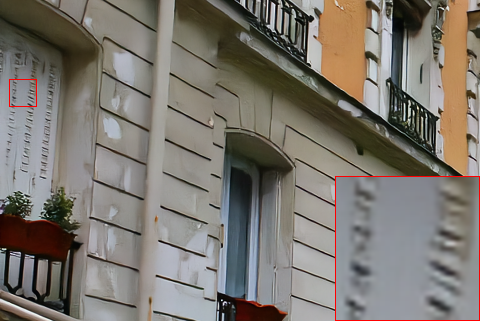} &
        \includegraphics[width=0.18\linewidth]{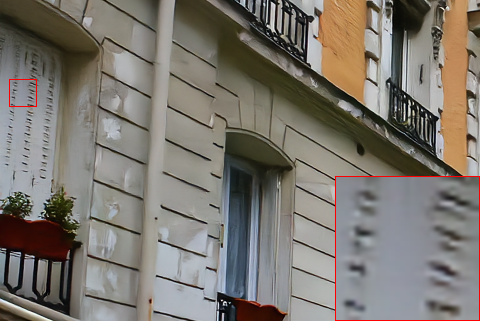} &
        \includegraphics[width=0.18\linewidth]{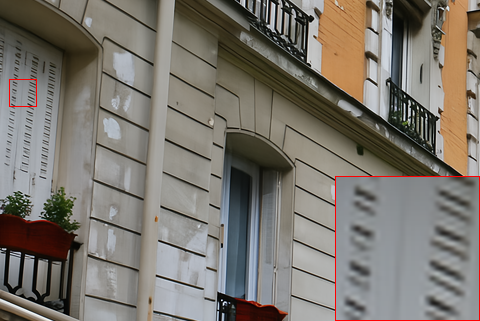} &
        \includegraphics[width=0.18\linewidth]{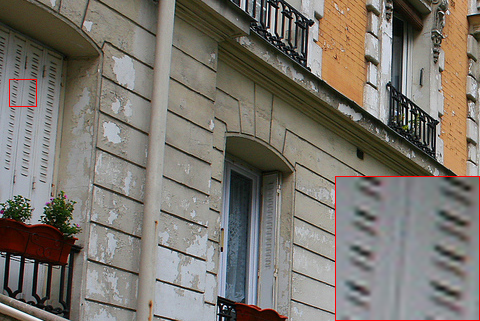}\\
        &\includegraphics[width=0.18\linewidth]{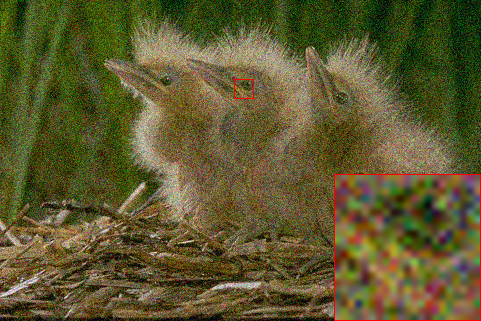} &
        \includegraphics[width=0.18\linewidth]{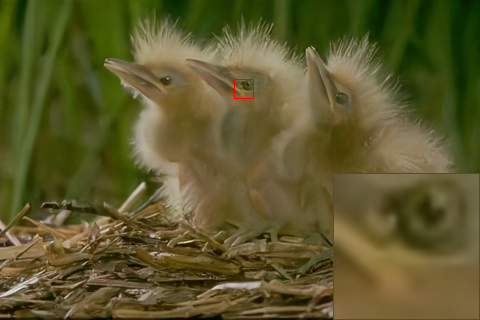} &
        \includegraphics[width=0.18\linewidth]{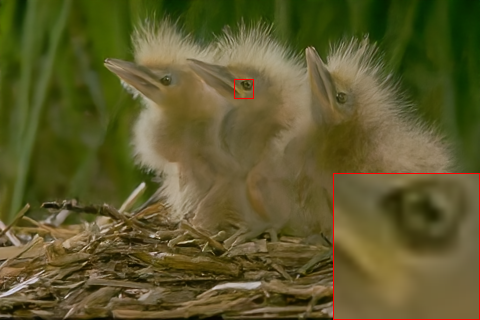} &
        \includegraphics[width=0.18\linewidth]{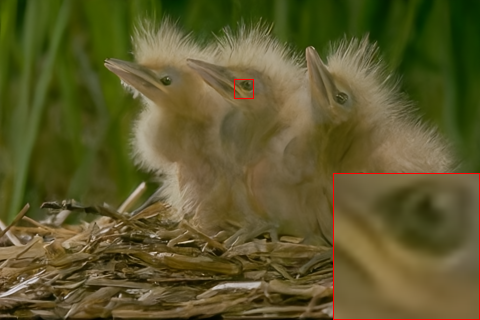} &
        \includegraphics[width=0.18\linewidth]{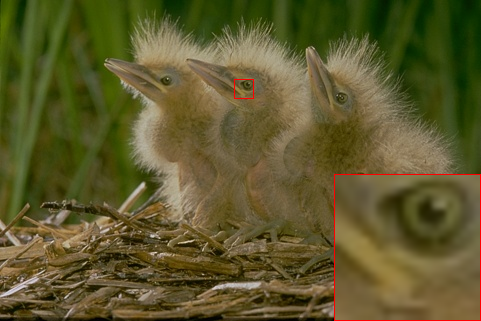}\\
         &Degraded & PromptIR & InstructIR & \textbf{VLU-Net (Ours)} & GT\\
    \end{tabular}}
    \vspace{-3mm}
    \caption{Visual comparison with state-of-the-art under NHR settings.}
    \label{fig:NHR}
    \vspace{-4mm}
\end{figure*}

\noindent \textbf{All-in-one Image Restoration Results.}
Table \ref{NHRBL} and \ref{NHR} present the results of VLU-Net compared with other state-of-the-art one-by-one or all-in-one approaches under NHRBL and NHR settings.
Figure \ref{fig:NHR} shows visual results of VLU-Net and other methods.
Our method exceeds InstructIR \cite{conde2024high} (all-in-one end-to-end method) by 0.56 dB on average, especially in dehazing and deraining, by 3.74 dB and 1.70 dB under NHRBL setting.
Compared to DGUNet (one-by-one DUN method) which also has interpretability as our VLU-Net, ours not only conducts restoration for multiple degradation tasks within single training process, but also achieves better results with 1.79 dB on average.
For NHR setting, VLU-Net achieves 0.27 dB average improvement, 0.49 dB for dehazing and 0.95 dB for deraining.

\begin{table*}[!t]
\centering
\setlength{\tabcolsep}{4pt}
\caption{Comparison of dehazing, deraining, and denoising results at noise levels $\sigma \in \{15, 25, 50\}$ under NHR settings. The PSNR (highlighted in orange cell) and SSIM metrics are reported, with the highest values indicated in \textcolor{myred}{Red} and the second highest in \textcolor{myblue}{Blue}.}
\vspace{-3mm}
\scalebox{0.8}{
\begin{tabular}{clll>{\columncolor{lightpeach!50}}cc>{\columncolor{lightpeach!50}}cc>{\columncolor{lightpeach!50}}cc>{\columncolor{lightpeach!50}}cc>{\columncolor{lightpeach!50}}cc>{\columncolor{lightpeach!50}}cc>{\columncolor{lightpeach!50}}c}
\toprule
\multirow{2}{*}{\text{Type}} & \multirow{2}{*}{\text{Method}} & \multirow{2}{*}{\text{Reference}} & \multirow{2}{*}{\text{Params.}} & \multicolumn{2}{c}{\text{Dehazing}} & \multicolumn{2}{c}{\text{Deraining}} &\multicolumn{6}{c}{\text{Denoising}} & \multicolumn{2}{c}{\text{Average}} \\
\cmidrule(l{0.7em}r{0.7em}){5-6}
\cmidrule(l{0.7em}r{0.7em}){7-8}
\cmidrule(l{0.7em}r{0.7em}){9-14}
\cmidrule(l{0.7em}r{0.7em}){15-16}
& & & &\multicolumn{2}{c}{\textit{SOTS}} & \multicolumn{2}{c}{\textit{Rain100L}} & \multicolumn{2}{c}{\textit{CBSD68}$_{\sigma=15}$}&\multicolumn{2}{c}{\textit{CBSD68}$_{\sigma=25}$} & \multicolumn{2}{c}{\textit{CBSD68}$_{\sigma=50}$} & \multicolumn{1}{c}{\text{PSNR}} & \multicolumn{1}{c}{\text{SSIM}}\\
\toprule
\multirow{3}{*}{\makecell{\text{One-by-one} \\ (end-to-end)} } & MPRNet \cite{zamir2021multi} & CVPR'21 & 16M & 28.00 & .958 & 33.86 & .958 & 33.27 & .920 & 30.76 & \text{.871} & \text{27.29} & .761 & 30.63&.894\\
 &Restormer \cite{zamir2022restormer} & CVPR'22 & 26M &27.78&.958& 33.78&.958& 33.72&.930& 30.67&.865 &27.63&.792&30.75&.901\\
 & MambaIR \cite{guo2024mambair} & ECCV'24& 27M & 29.57&.970& 35.42&.969&33.88&.931 &30.95&.874 &27.74&.793&31.51&.907\\
\toprule
\multirow{7}{*}{\makecell{All-in-one \\ (end-to-end)}} & AirNet \cite{li2022all} & CVPR'22& 9M &27.94&.962 &34.90&.967& 33.92&.932 &31.26& .888 &28.00&.797&31.20&.910\\
 &IDR \cite{zhang2023ingredient} &CVPR'23& 15M &29.87&.970& 36.03&.971&33.89&.931 &31.32&.884& 28.04&.798&31.83&.911\\
 &PromptIR \cite{potlapalli2306promptir} &NeurIPS'23& 33M & \textcolor{myblue}{\textbf{30.58}}& \textcolor{myblue}{\textbf{.974}} &36.37&.972&33.98& \textcolor{myblue}{\textbf{.933}}& 31.31& .888 &28.06&.799&32.06&\textcolor{myblue}{\textbf{.913}}\\
 &NDR \cite{yao2024neural} &TIP'24& 39M & 28.64&.962& 35.42&.969&34.01&.932 &31.36&.887& 28.10&.798& 31.51&.910\\
 &Gridformer \cite{wang2024gridformer} &IJCV’24& 34M &30.37&.970 &37.15&.972&33.93&.931& 31.37&.887 &28.11& \textcolor{myblue}{\textbf{.801}}&32.19&.912\\
 &InstructIR \cite{conde2024high} &ECCV'24& 16M & 30.22 & .959& \textcolor{myblue}{\textbf{37.98}}& \textcolor{myblue}{\textbf{.978}}& \textcolor{myred}{\textbf{34.15}}& \textcolor{myblue}{\textbf{.933}}& \textcolor{myred}{\textbf{31.52}}& \textcolor{myblue}{\textbf{.890}}& \textcolor{myred}{\textbf{28.30}}& \textcolor{myred}{\textbf{.804}}&\textcolor{myblue}{\textbf{32.43}}&\textcolor{myblue}{\textbf{.913}}\\
 \cmidrule(l{0.7em}r{0.7em}){2-16}
  \rowcolor{lightgray!20}
  \cellcolor{white} (deep unfolding) & VLU-Net & Ours & 35M &  \textcolor{myred}{\textbf{30.71}} &  \textcolor{myred}{\textbf{.980}} &  \textcolor{myred}{\textbf{38.93}} &  \textcolor{myred}{\textbf{.984}} &  \textcolor{myblue}{\textbf{34.13}} &  \textcolor{myred}{\textbf{.935}} &  \textcolor{myblue}{\textbf{31.48}} &  \textcolor{myred}{\textbf{.892}} &  \textcolor{myblue}{\textbf{28.23}} &  \textcolor{myred}{\textbf{.804}} &  \textcolor{myred}{\textbf{32.70}} &  \textcolor{myred}{\textbf{.919}}\\
\bottomrule
\end{tabular}
}
\label{NHR}
\vspace{-4mm}
\end{table*}

\noindent \textbf{One-by-one Image Restoration Results.}
We compare our VLU-Net for three tasks, denoising, dehazing and deraining in Table \ref{tab:noise} and \ref{tab:hazeandrain}.
We achieve 0.90 dB average improvement for denoising in \textit{Urban100} dataset, 1.21 dB for dehazing and 0.62 dB for deraining compared to deep end-to-end learning method InstructIR \cite{conde2024high}.
Compared to current one-by-one DUN method DGUNet \cite{mou2022deep}, our all-in-one VLU-Net not only expands the DUN architecture with hierarchical feature level information extraction, unleashing the potential degradation analysis capabilities of GDM, but also achieves 1.00 dB average improvement in denoising \textit{Urban100} and \textit{CBSD68}, 1.18 dB for deraining \textit{Rain100L}.

\begin{table}
    \centering
\setlength{\tabcolsep}{3.2pt}
    \caption{Results of one-by-one IR for $\sigma \in \{15,25,50\}$ denoising.}
    \vspace{-3mm}
\scalebox{0.68}{
    \begin{tabular}{lcccccc}
    \toprule
        \multirow{2}{*}{\text{Method}} & \multicolumn{3}{c}{\textit{CBSD68}} & \multicolumn{3}{c}{\textit{Urban100}} \\
        \cmidrule(l{0.7em}r{0.7em}){2-4}
        \cmidrule(l{0.7em}r{0.7em}){5-7}
        &$\sigma=15$&$\sigma=25$&$\sigma=50$&$\sigma=15$&$\sigma=25$&$\sigma=50$\\
    \toprule
        (TIP'17) DnCNN \cite{zhang2017beyond} &33.90&31.24&27.95&32.98&30.81&27.59\\
        (TIP'18) FFDNet \cite{zhang2018ffdnet} &33.87&31.21&27.96&33.83&31.40&28.05\\
        (ICCV'21) SwinIR \cite{liang2021swinir} &33.31&30.59&27.13&32.79&30.18&26.52\\
        (CVPR'22) DGUNet \cite{mou2022deep} &33.85&31.10&27.92&33.67&31.27&27.94\\
        (CVPR'22) Restormer \cite{zamir2022restormer} &34.03&31.49&28.11&33.72&31.26&28.03\\
    \toprule
        (CVPR'22) AirNet \cite{li2022all} &34.14&31.48&28.23&34.40&32.10&28.88\\
        (NeurIPS'23) PromptIR \cite{potlapalli2306promptir} &\textcolor{myblue}{\textbf{34.34}}&\textcolor{myblue}{\textbf{31.71}}&\textcolor{myred}{\textbf{28.49}}&\textcolor{myblue}{\textbf{34.77}}&\textcolor{myblue}{\textbf{32.49}}&\textcolor{myblue}{\textbf{29.39}}\\
        (CVPR'23) IDR \cite{zhang2023ingredient} & 34.11 &31.60 &28.14& 33.82 &31.29 &28.07\\
        (ECCV'24) InstructIR \cite{conde2024high} &34.15&31.52& 28.30& 34.12& 31.80& 28.63\\
        \rowcolor{lightgray!20}
        VLU-Net (ours) & \textcolor{myred}{\textbf{34.35}} & \textcolor{myred}{\textbf{31.72}} & \textcolor{myblue}{\textbf{28.46}} & \textcolor{myred}{\textbf{34.92}} &\textcolor{myred}{\textbf{32.71}}&\textcolor{myred}{\textbf{29.61}}\\
    \bottomrule
    \end{tabular}}
    \label{tab:noise}
    \vspace{-2mm}
\end{table}

\begin{table}
    \centering
\setlength{\tabcolsep}{3.2pt}
    \caption{Results of one-by-one IR for dehazing and deraining.}
    \vspace{-3mm}
\scalebox{0.66}{
    \begin{tabular}{l>{\columncolor{lightpeach!50}}clc>{\columncolor{lightpeach!50}}cc}
    \toprule
        \multirow{2}{*}{\text{Method}} & \multicolumn{2}{c}{\text{Dehazing}} & \multirow{2}{*}{\text{Method}} & \multicolumn{2}{c}{\text{Deraining}} \\
        \cmidrule(l{0.7em}r{0.7em}){2-3}
        \cmidrule(l{0.7em}r{0.7em}){5-6}
        &\multicolumn{2}{c}{\textit{SOTS}}& &\multicolumn{2}{c}{\textit{Rain100L}}\\
    \toprule
    (TIP'16) DehazeNet\cite{cai2016dehazenet}&22.46&.851&(CVPR'19) UMR\cite{yasarla2019uncertainty}&32.39&.921\\
       (CVPR'19) EPDN \cite{qu2019enhanced} &22.57&.863&(CVPR'19) SIRR\cite{wei2019semi}&32.37&.926\\
        (AAAI'20) FDGAN \cite{dong2020fd} &23.15&.921&(CVPR'20) MSPFN \cite{jiang2020multi}&33.50&.948\\
        (PAMI'23) FSNet \cite{cui2023image} &31.11&.971&(CVPR'22) DGUNet \cite{mou2022deep}&37.42&.969\\
        (CVPR'22) Restormer \cite{zamir2022restormer} &30.87&.969&(CVPR'22) Restormer \cite{zamir2022restormer}&36.74&.978\\
    \toprule
       (CVPR’22) AirNet \cite{li2022all} &23.18&.900&(CVPR’22) AirNet \cite{li2022all}&34.90&.977\\
        (NeurIPS'23) PromptIR \cite{potlapalli2306promptir} &\textcolor{myblue}{\textbf{31.31}}&\textcolor{myblue}{\textbf{.973}}&(NeurIPS'23) PromptIR \cite{potlapalli2306promptir}&37.04&\textcolor{myblue}{\textbf{.979}}\\
        (ECCV'24) InstructIR \cite{conde2024high} &30.22&.959&(ECCV'24) InstructIR \cite{conde2024high}&\textcolor{myblue}{\textbf{37.98}}&.978\\
    \rowcolor{lightgray!20}
        VLU-Net (ours) & \textcolor{myred}{\textbf{31.43}} &\textcolor{myred}{\textbf{.980}} &VLU-Net (ours)& \textcolor{myred}{\textbf{38.60}} &\textcolor{myred}{\textbf{.984}}\\
    \bottomrule
    \end{tabular}}
    \label{tab:hazeandrain}
    \vspace{-1mm}
\end{table}

\subsection{Ablation Analysis} \label{sec:abalation}
\begin{figure}
    \centering
    \includegraphics[width=0.23\linewidth]{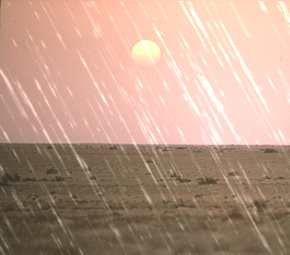}
    \includegraphics[width=0.23\linewidth]{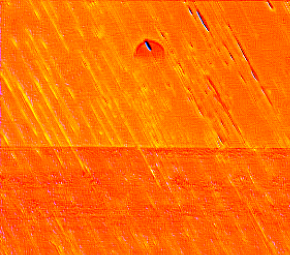}
    \includegraphics[width=0.23\linewidth]{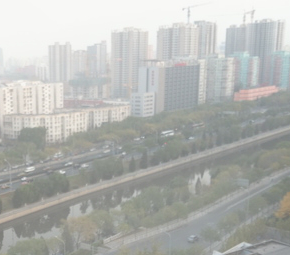}
    \includegraphics[width=0.23\linewidth]{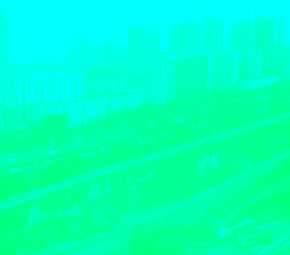}\\
    \vspace{-2mm}
    \caption{Visual degradation from D-GDM for rain and haze.}
    \label{fig:visual_degradation}
    \vspace{-3mm}
\end{figure}

\vspace{-1mm}
\noindent \textbf{Visual Effect of D-GDM.}
We present the Figure \ref{fig:visual_degradation}, the visual results of the degradation map generated by D-GDM for rain and haze.
Figure \ref{fig:visual_degradation} shows that the proposed D-GDM generates different degradation for different types of inputs, which verifies the clarification abilities of D-GDM.

\begin{figure}
    \centering
    \begin{tabular}{c}
         \includegraphics[width=1\linewidth]{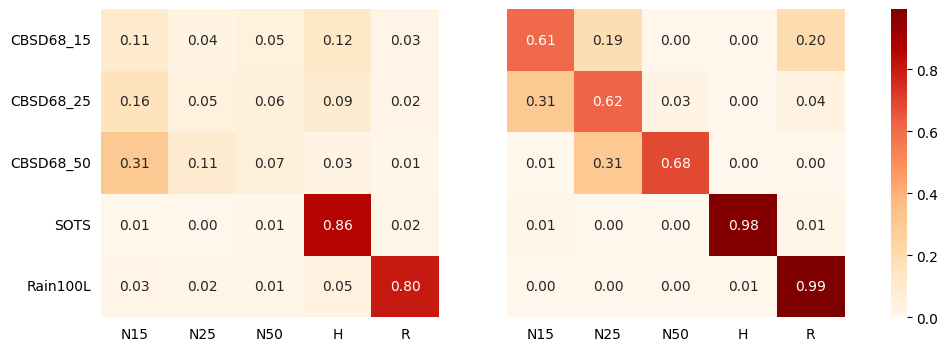}\\
          \includegraphics[width=1\linewidth]{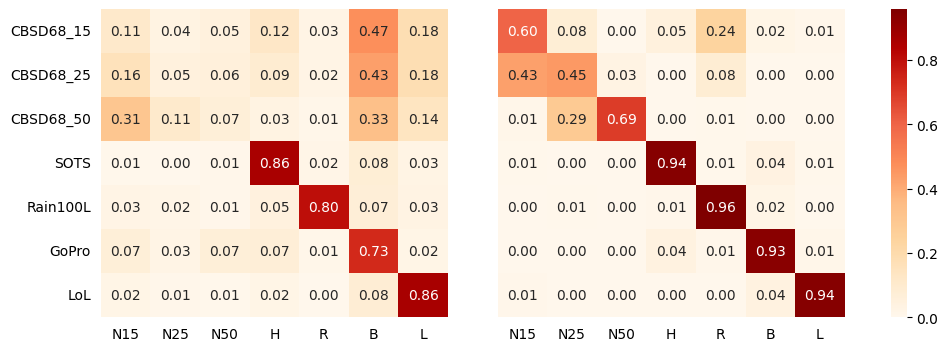}
    \end{tabular}
    \vspace{-4mm}
    \caption{Heat-maps for CLIP before (Left) and after (Right) fine-tuning. Top is for NHR setting and bottom is for NHRBL setting.}
    \label{fig:clip_heatmaps}
    \vspace{-3mm}
\end{figure}

\noindent \textbf{Effect of Fine-tuned CLIP.}
We verify the effect of fine-tuned CLIP from two aspects.
Firstly, as stated in Sec. \ref{sec:motivation}, we provide the results about the similarities we tested for CLIP and the degradation fine-tuned CLIP in Figure \ref{fig:clip_heatmaps}.
For each degradation type, we average the similarity scores within each test dataset.
Each row represents the similarity of target test dataset relative to each degradation type.
After fine-tuning CLIP shows more powerful classification capabilities for different degradation types, especially for the noise detection with a step up from nothing, also $12\%$ and $19\%$ improvement for rain and haze.
Secondly, we present the all-in-one IR results of ``w/o CLIP'' in Table \ref{tab:ablation_NHRBL}.
Ours achieves 0.3 dB improvement against ``w/o CLIP''.

\begin{table}
    \centering
\setlength{\tabcolsep}{3.2pt}
    \caption{Ablation IR results under NHRBL setting.}
    \vspace{-3mm}
\scalebox{0.7}{
    \begin{tabular}{l>{\columncolor{lightpeach!50}}cl>{\columncolor{lightpeach!50}}cc>{\columncolor{lightpeach!50}}cc>{\columncolor{lightpeach!50}}cc}
    \toprule
        \text{Method} & \multicolumn{2}{c}{\text{Dehazing}} & \multicolumn{2}{c}{\text{Denoising}} & \multicolumn{2}{c}{\text{Low-light}} & \multicolumn{2}{c}{\text{Average}}\\
    \toprule
    End-to-End Transformer~\cite{zamir2022restormer}&24.09&.927&\textcolor{myred}{\textbf{31.49}}&.884&20.41&.806&25.33&.872\\
    Previous DUN~\cite{mou2022deep}&24.78&.940&31.10&.883&21.87&\textcolor{myblue}{\textbf{.823}}&25.92&.882\\
    Hierarchical w/o CLIP &30.24&.968&31.40&.886&22.02&.819&27.89&.891\\
    Hierarchical w/ CLIP &\textcolor{myblue}{\textbf{30.44}}&\textcolor{myblue}{\textbf{.974}}&31.41&\textcolor{myblue}{\textbf{.889}}&\textcolor{myblue}{\textbf{22.12}}&.822&\textcolor{myblue}{\textbf{27.97}}&\textcolor{myblue}{\textbf{.895}}\\
    \rowcolor{lightgray!20}
    Hierarchical w/ tunedCLIP & \textcolor{myred}{\textbf{30.84}} &\textcolor{myred}{\textbf{.980}}& \textcolor{myblue}{\textbf{31.43}} &\textcolor{myred}{\textbf{.891}}&\textcolor{myred}{\textbf{22.29}} & \textcolor{myred}{\textbf{.833}}&\textcolor{myred}{\textbf{28.19}}&\textcolor{myred}{\textbf{.901}}\\
    \bottomrule
    \end{tabular}}
    \label{tab:ablation_NHRBL}
    \vspace{-2mm}
\end{table}

\noindent \textbf{Effect of the Hierarchical DUN.}
Besides the explanation in Sec. \ref{sec:HDUN}, We also verify the effect of the hierarchical DUN from below two aspects: firstly we present the architecture comparison figure between current DUN and our hierarchical DUN in the supplementary materials for better framework clarification, secondly we compare the results under NHRBL setting in Table \ref{tab:ablation_NHRBL} among our VLU-Net, the w/ CLIP version, the w/o CLIP version, end-to-end deep learning method Restormer \cite{zamir2022restormer} and DUN method DGUNet \cite{mou2022deep}.
According to the results of w/o CLIP version in Table \ref{tab:ablation_NHRBL}, our basic hierarchical DUN (w/o CLIP) also achieve comparable results with state-of-the-art and is higher than DGUNet 1.97 dB on average in PSNR, also 2.56 dB for Restormer.
It shows that the feature level unfolding releases the potential of the DGM in DUN, facilitating efficient information transmission.
Also it promotes multi-level features without the compression-decompression constrains between stages, leveraging different hierarchical features for stages with different responsibilities.

\noindent \textbf{Denoising results under random generated noise levels.}
To verify the unseen degradation removal performance of our proposed VLU-Net, we present the denoising results under random generated noise. 
We average the results within each dataset and each image has the random generated noise level $\sigma \in [15, 50]$ tested by the pre-trained VLU-Net under NHR setting.
From Table \ref{tab:random_denoising}, our VLU-Net achieves 0.39 dB improvement in \textit{Urban100} dataset.

\begin{table}
\vspace{-1mm}
    \centering
\setlength{\tabcolsep}{3.2pt}
    \caption{Results under random generated $\sigma \in [15, 50]$ noise.}
    \vspace{-3mm}
\scalebox{0.8}{
    \begin{tabular}{l>{\columncolor{lightpeach!50}}cl>{\columncolor{lightpeach!50}}cc>{\columncolor{lightpeach!50}}cc}
    \toprule
        \text{Method} & \multicolumn{2}{c}{\textit{CBSD68}} & \multicolumn{2}{c}{\textit{Urban100}} & \multicolumn{2}{c}{\text{Average}}\\
    \toprule
    (NeurIPS'23) PromptIR\cite{potlapalli2306promptir}&30.43&.863&30.58&.897&30.51&.880\\
    \rowcolor{lightgray!20}
        VLU-Net (ours) & \textcolor{myred}{\textbf{30.53}} &\textcolor{myred}{\textbf{.864}}& \textcolor{myred}{\textbf{30.97}} &\textcolor{myred}{\textbf{.901}}& \textcolor{myred}{\textbf{30.75}} &\textcolor{myred}{\textbf{.883}}\\
    \bottomrule
    \end{tabular}}
    \label{tab:random_denoising}
    \vspace{-5mm}
\end{table}

\vspace{-1mm}
\section{Conclusion}
\vspace{-1mm}
We addressed the critical challenges posed by dynamic image degradations, often caused by sensor limitations or adverse environmental conditions. 
Existing deep unfolding networks demonstrated stable restoration performance but were constrained by the manual selection of degradation transforms, limiting the adaptability across diverse scenarios. 
To overcome these limitations, we proposed VLU-Net, which employed a unified framework capable of handling multiple degradation types simultaneously by leveraging a refined vision-language model for automatic feature alignment and transformation selection. 
By enhancing adaptability and interpretability, VLU-Net established a hierarchical framework for multi-level information integration, outperforming leading one-by-one or all-in-one methods.
This work opened pathways for more flexible, efficient and robust solutions in complex, real-world imaging scenarios.

\section{Acknowledgments}
This work was supported  by National Natural Science Foundation of China (grant No. 62350710797) and the Key Program of Technology Research from Shenzhen Science and Technology Innovation Committee under Grant JSGG20220831104402004.
{
    \small
    \bibliographystyle{ieeenat_fullname}
    \bibliography{main}

\begin{thebibliography}{55}
\providecommand{\natexlab}[1]{#1}
\providecommand{\url}[1]{\texttt{#1}}
\expandafter\ifx\csname urlstyle\endcsname\relax
  \providecommand{\doi}[1]{doi: #1}\else
  \providecommand{\doi}{doi: \begingroup \urlstyle{rm}\Url}\fi

\bibitem[Beck and Teboulle(2009)]{beck2009fast}
Amir Beck and Marc Teboulle.
\newblock A fast iterative shrinkage-thresholding algorithm for linear inverse problems.
\newblock \emph{SIAM Journal on Imaging Sciences}, 2\penalty0 (1):\penalty0 183--202, 2009.

\bibitem[Bertero et~al.(2021)Bertero, Boccacci, and De~Mol]{bertero2021introduction}
Mario Bertero, Patrizia Boccacci, and Christine De~Mol.
\newblock \emph{Introduction to inverse problems in imaging}.
\newblock CRC press, 2021.

\bibitem[Bioucas-Dias and Figueiredo.(2007)]{twist}
J.M. Bioucas-Dias and M.A.T. Figueiredo.
\newblock A new twist: Two-step iterative shrinkage/thresholding algorithms for image restoration.
\newblock \emph{IEEE TIP}, 2007.

\bibitem[Cai et~al.(2016)Cai, Xu, Jia, Qing, and Tao]{cai2016dehazenet}
Bolun Cai, Xiangmin Xu, Kui Jia, Chunmei Qing, and Dacheng Tao.
\newblock Dehazenet: An end-to-end system for single image haze removal.
\newblock \emph{IEEE TIP}, 25\penalty0 (11):\penalty0 5187--5198, 2016.

\bibitem[Cai et~al.(2022{\natexlab{a}})Cai, Lin, Hu, Wang, Yuan, Zhang, Timofte, and Gool]{mst}
Yuanhao Cai, Jing Lin, Xiaowan Hu, Haoqian Wang, Xin Yuan, Yulun Zhang, Radu Timofte, and Luc~Van Gool.
\newblock Mask-guided spectral-wise transformer for efficient hyperspectral image reconstruction.
\newblock In \emph{CVPR}, 2022{\natexlab{a}}.

\bibitem[Cai et~al.(2022{\natexlab{b}})Cai, Lin, Hu, Wang, Yuan, Zhang, Timofte, and Van~Gool]{cst}
Yuanhao Cai, Jing Lin, Xiaowan Hu, Haoqian Wang, Xin Yuan, Yulun Zhang, Radu Timofte, and Luc Van~Gool.
\newblock Coarse-to-fine sparse transformer for hyperspectral image reconstruction.
\newblock In \emph{ECCV}, pages 686--704. Springer, 2022{\natexlab{b}}.

\bibitem[Cao et~al.(2024)Cao, Cao, Pang, Meng, and Cao]{cao2024hair}
Jin Cao, Yi Cao, Li Pang, Deyu Meng, and Xiangyong Cao.
\newblock Hair: Hypernetworks-based all-in-one image restoration.
\newblock \emph{arXiv preprint arXiv:2408.08091}, 2024.

\bibitem[Chang et~al.(2020)Chang, Yan, Zhao, Fang, Zhang, and Zhong]{chang2020weighted}
Yi Chang, Luxin Yan, Xi-Le Zhao, Houzhang Fang, Zhijun Zhang, and Sheng Zhong.
\newblock Weighted low-rank tensor recovery for hyperspectral image restoration.
\newblock \emph{IEEE TCYB}, 50\penalty0 (11):\penalty0 4558--4572, 2020.

\bibitem[Chen et~al.(2017)Chen, Guo, Wang, Wang, Peng, and He]{NonLRMA}
Yongyong Chen, Yanwen Guo, Yongli Wang, Dong Wang, Chong Peng, and Guoping He.
\newblock Denoising of hyperspectral images using nonconvex low rank matrix approximation.
\newblock \emph{IEEE TGRS}, 55\penalty0 (9):\penalty0 5366--5380, 2017.

\bibitem[Chen et~al.(2024)Chen, Zhang, Liu, Gu, Kong, Yuan, et~al.]{chen2024hierarchical}
Zheng Chen, Yulun Zhang, Ding Liu, Jinjin Gu, Linghe Kong, Xin Yuan, et~al.
\newblock Hierarchical integration diffusion model for realistic image deblurring.
\newblock In \emph{NeurIPS}, 2024.

\bibitem[Conde et~al.(2024)Conde, Geigle, and Timofte]{conde2024high}
Marcos~V Conde, Gregor Geigle, and Radu Timofte.
\newblock Instructir: High-quality image restoration following human instructions.
\newblock In \emph{ECCV}, 2024.

\bibitem[Cui et~al.(2023)Cui, Ren, Cao, and Knoll]{cui2023image}
Yuning Cui, Wenqi Ren, Xiaochun Cao, and Alois Knoll.
\newblock Image restoration via frequency selection.
\newblock \emph{IEEE TPAMI}, 2023.

\bibitem[Dong et~al.(2020)Dong, Liu, Zhang, Chen, and Qiao]{dong2020fd}
Yu Dong, Yihao Liu, He Zhang, Shifeng Chen, and Yu Qiao.
\newblock Fd-gan: Generative adversarial networks with fusion-discriminator for single image dehazing.
\newblock In \emph{AAAI}, pages 10729--10736, 2020.

\bibitem[Guo et~al.(2024)Guo, Li, Dai, Ouyang, Ren, and Xia]{guo2024mambair}
Hang Guo, Jinmin Li, Tao Dai, Zhihao Ouyang, Xudong Ren, and Shu-Tao Xia.
\newblock Mambair: A simple baseline for image restoration with state-space model.
\newblock In \emph{ECCV}, 2024.

\bibitem[Guo et~al.(2023)Guo, Wang, Yang, Huang, Wang, Pfister, and Wen]{guo2023shadowdiffusion}
Lanqing Guo, Chong Wang, Wenhan Yang, Siyu Huang, Yufei Wang, Hanspeter Pfister, and Bihan Wen.
\newblock Shadowdiffusion: When degradation prior meets diffusion model for shadow removal.
\newblock In \emph{CVPR}, pages 14049--14058, 2023.

\bibitem[Hendrycks and Gimpel(2016)]{hendrycks2016gelu}
Dan Hendrycks and Kevin Gimpel.
\newblock Gaussian error linear units (gelus).
\newblock \emph{arXiv preprint arXiv:1606.08415}, 2016.

\bibitem[Hyun~Kim et~al.(2017)Hyun~Kim, Mu~Lee, Scholkopf, and Hirsch]{hyun2017online}
Tae Hyun~Kim, Kyoung Mu~Lee, Bernhard Scholkopf, and Michael Hirsch.
\newblock Online video deblurring via dynamic temporal blending network.
\newblock In \emph{ICCV}, 2017.

\bibitem[Jia et~al.(2021)Jia, Yang, Xia, Chen, Parekh, Pham, Le, Sung, Li, and Duerig]{jia2021scaling}
Chao Jia, Yinfei Yang, Ye Xia, Yi-Ting Chen, Zarana Parekh, Hieu Pham, Quoc Le, Yun-Hsuan Sung, Zhen Li, and Tom Duerig.
\newblock Scaling up visual and vision-language representation learning with noisy text supervision.
\newblock In \emph{ICML}, pages 4904--4916. PMLR, 2021.

\bibitem[Jiang et~al.(2020)Jiang, Wang, Yi, Chen, Huang, Luo, Ma, and Jiang]{jiang2020multi}
Kui Jiang, Zhongyuan Wang, Peng Yi, Chen Chen, Baojin Huang, Yimin Luo, Jiayi Ma, and Junjun Jiang.
\newblock Multi-scale progressive fusion network for single image deraining.
\newblock In \emph{CVPR}, pages 8346--8355, 2020.

\bibitem[Kawar et~al.(2022)Kawar, Elad, Ermon, and Song]{ddrm}
Bahjat Kawar, Michael Elad, Stefano Ermon, and Jiaming Song.
\newblock Denoising diffusion restoration models.
\newblock In \emph{NeurIPS}, pages 23593--23606, 2022.

\bibitem[Kolda and Bader(2009)]{kolda2009tensor}
Tamara~G Kolda and Brett~W Bader.
\newblock Tensor decompositions and applications.
\newblock \emph{SIAM Review}, 51\penalty0 (3):\penalty0 455--500, 2009.

\bibitem[Li et~al.(2022{\natexlab{a}})Li, Liu, Hu, Wu, Lv, and Peng]{li2022all}
Boyun Li, Xiao Liu, Peng Hu, Zhongqin Wu, Jiancheng Lv, and Xi Peng.
\newblock All-in-one image restoration for unknown corruption.
\newblock In \emph{CVPR}, pages 17452--17462, 2022{\natexlab{a}}.

\bibitem[Li et~al.(2022{\natexlab{b}})Li, Li, Xiong, and Hoi]{li2022blip}
Junnan Li, Dongxu Li, Caiming Xiong, and Steven Hoi.
\newblock Blip: Bootstrapping language-image pre-training for unified vision-language understanding and generation.
\newblock In \emph{ICML}, pages 12888--12900. PMLR, 2022{\natexlab{b}}.

\bibitem[Liang et~al.(2021)Liang, Cao, Sun, Zhang, Van~Gool, and Timofte]{liang2021swinir}
Jingyun Liang, Jiezhang Cao, Guolei Sun, Kai Zhang, Luc Van~Gool, and Radu Timofte.
\newblock Swin\protect{IR}: Image restoration using swin transformer.
\newblock In \emph{ICCV}, pages 1833--1844, 2021.

\bibitem[Luo et~al.(2024{\natexlab{a}})Luo, Gustafsson, Zhao, Sj{\"o}lund, and Sch{\"o}n]{luo2024controlling}
Ziwei Luo, Fredrik~K Gustafsson, Zheng Zhao, Jens Sj{\"o}lund, and Thomas~B Sch{\"o}n.
\newblock Controlling vision-language models for multi-task image restoration.
\newblock In \emph{ICLR}, 2024{\natexlab{a}}.

\bibitem[Luo et~al.(2024{\natexlab{b}})Luo, Gustafsson, Zhao, Sj{\"o}lund, and Sch{\"o}n]{luo2024photo}
Ziwei Luo, Fredrik~K Gustafsson, Zheng Zhao, Jens Sj{\"o}lund, and Thomas~B Sch{\"o}n.
\newblock Photo-realistic image restoration in the wild with controlled vision-language models.
\newblock In \emph{CVPR}, pages 6641--6651, 2024{\natexlab{b}}.

\bibitem[Ma et~al.(2019)Ma, Liu, Shou, and Yuan]{admm-net}
Jiawei Ma, Xiao-Yang Liu, Zheng Shou, and Xin Yuan.
\newblock Deep tensor \protect{ADMM}-net for snapshot compressive imaging.
\newblock In \emph{ICCV}, 2019.

\bibitem[Meng et~al.(2023)Meng, Yuan, and Jalali]{meng2023deep}
Ziyi Meng, Xin Yuan, and Shirin Jalali.
\newblock Deep unfolding for snapshot compressive imaging.
\newblock \emph{IJCV}, 131\penalty0 (11):\penalty0 2933--2958, 2023.

\bibitem[Miao et~al.(2023)Miao, Zhang, Zhang, and Tao]{dds2m}
Yuchun Miao, Lefei Zhang, Liangpei Zhang, and Dacheng Tao.
\newblock Dds2m: Self-supervised denoising diffusion spatio-spectral model for hyperspectral image restoration.
\newblock In \emph{ICCV}, pages 12086--12096, 2023.

\bibitem[Mou et~al.(2022)Mou, Wang, and Zhang]{mou2022deep}
Chong Mou, Qian Wang, and Jian Zhang.
\newblock Deep generalized unfolding networks for image restoration.
\newblock In \emph{CVPR}, pages 17399--17410, 2022.

\bibitem[Pan et~al.(2020)Pan, Bai, Tang, and b]{tsp}
Jinshan Pan, Haoran Bai, Jinhui Tang, and b.
\newblock Cascaded deep video deblurring using temporal sharpness prior.
\newblock In \emph{CVPR}, 2020.

\bibitem[Peng et~al.(2020)Peng, Xie, Zhao, Wang, Yee, and Meng]{E3DTV}
Jiangjun Peng, Qi Xie, Qian Zhao, Yao Wang, Leung Yee, and Deyu Meng.
\newblock Enhanced 3dtv regularization and its applications on \protect{HSI} denoising and compressed sensing.
\newblock \emph{IEEE TIP}, 29:\penalty0 7889--7903, 2020.

\bibitem[Potlapalli et~al.(2023)Potlapalli, Zamir, Khan, and Khan]{potlapalli2306promptir}
Vaishnav Potlapalli, Syed~Waqas Zamir, Salman Khan, and Fahad Khan.
\newblock Promptir: Prompting for all-in-one image restoration.
\newblock In \emph{NeurIPS}, 2023.

\bibitem[Qu et~al.(2019)Qu, Chen, Huang, and Xie]{qu2019enhanced}
Yanyun Qu, Yizi Chen, Jingying Huang, and Yuan Xie.
\newblock Enhanced pix2pix dehazing network.
\newblock In \emph{CVPR}, pages 8160--8168, 2019.

\bibitem[Radford et~al.(2021)Radford, Kim, Hallacy, Ramesh, Goh, Agarwal, Sastry, Askell, Mishkin, Clark, et~al.]{radford2021learning}
Alec Radford, Jong~Wook Kim, Chris Hallacy, Aditya Ramesh, Gabriel Goh, Sandhini Agarwal, Girish Sastry, Amanda Askell, Pamela Mishkin, Jack Clark, et~al.
\newblock Learning transferable visual models from natural language supervision.
\newblock In \emph{ICML}, pages 8748--8763. PMLR, 2021.

\bibitem[Ryu et~al.(2019)Ryu, Liu, Wang, Chen, Wang, and Yin]{ryu2019plug}
Ernest Ryu, Jialin Liu, Sicheng Wang, Xiaohan Chen, Zhangyang Wang, and Wotao Yin.
\newblock Plug-and-play methods provably converge with properly trained denoisers.
\newblock In \emph{ICML}, pages 5546--5557. PMLR, 2019.

\bibitem[Valanarasu et~al.(2022)Valanarasu, Yasarla, and Patel]{valanarasu2022transweather}
Jeya Maria~Jose Valanarasu, Rajeev Yasarla, and Vishal~M Patel.
\newblock Transweather: Transformer-based restoration of images degraded by adverse weather conditions.
\newblock In \emph{CVPR}, pages 2353--2363, 2022.

\bibitem[Venkatakrishnan et~al.(2013)Venkatakrishnan, Bouman, and Wohlberg]{venkatakrishnan2013plug}
Singanallur~V Venkatakrishnan, Charles~A Bouman, and Brendt Wohlberg.
\newblock Plug-and-play priors for model based reconstruction.
\newblock In \emph{IEEE GlobalSIP}, pages 945--948, 2013.

\bibitem[Wang et~al.(2024)Wang, Zhang, Shao, Luo, Stenger, Lu, Kim, Liu, and Li]{wang2024gridformer}
Tao Wang, Kaihao Zhang, Ziqian Shao, Wenhan Luo, Bjorn Stenger, Tong Lu, Tae-Kyun Kim, Wei Liu, and Hongdong Li.
\newblock Gridformer: Residual dense transformer with grid structure for image restoration in adverse weather conditions.
\newblock \emph{IJCV}, pages 1--23, 2024.

\bibitem[Wang and Gan(2024)]{wang2024ufc}
Xiaoyang Wang and Hongping Gan.
\newblock Ufc-net: Unrolling fixed-point continuous network for deep compressive sensing.
\newblock In \emph{CVPR}, pages 25149--25159, 2024.

\bibitem[Wei et~al.(2019)Wei, Meng, Zhao, Xu, and Wu]{wei2019semi}
Wei Wei, Deyu Meng, Qian Zhao, Zongben Xu, and Ying Wu.
\newblock Semi-supervised transfer learning for image rain removal.
\newblock In \emph{CVPR}, pages 3877--3886, 2019.

\bibitem[Wu et~al.(2022)Wu, Weng, Zhang, Wang, Yang, and Jiang]{wu2022uretinex}
Wenhui Wu, Jian Weng, Pingping Zhang, Xu Wang, Wenhan Yang, and Jianmin Jiang.
\newblock Uretinex-net: Retinex-based deep unfolding network for low-light image enhancement.
\newblock In \emph{CVPR}, pages 5901--5910, 2022.

\bibitem[Yao et~al.(2024)Yao, Xu, Guan, Huang, and Xiong]{yao2024neural}
Mingde Yao, Ruikang Xu, Yuanshen Guan, Jie Huang, and Zhiwei Xiong.
\newblock Neural degradation representation learning for all-in-one image restoration.
\newblock \emph{IEEE TIP}, 2024.

\bibitem[Yasarla and Patel(2019)]{yasarla2019uncertainty}
Rajeev Yasarla and Vishal~M Patel.
\newblock Uncertainty guided multi-scale residual learning-using a cycle spinning cnn for single image de-raining.
\newblock In \emph{CVPR}, pages 8405--8414, 2019.

\bibitem[Yuan(2016)]{gap_tv}
Xin Yuan.
\newblock Generalized alternating projection based total variation minimization for compressive sensing.
\newblock In \emph{ICIP}, 2016.

\bibitem[Zamir et~al.(2021)Zamir, Arora, Khan, Hayat, Khan, Yang, and Shao]{zamir2021multi}
Syed~Waqas Zamir, Aditya Arora, Salman Khan, Munawar Hayat, Fahad~Shahbaz Khan, Ming-Hsuan Yang, and Ling Shao.
\newblock Multi-stage progressive image restoration.
\newblock In \emph{CVPR}, pages 14821--14831, 2021.

\bibitem[Zamir et~al.(2022)Zamir, Arora, Khan, Hayat, Khan, and Yang]{zamir2022restormer}
Syed~Waqas Zamir, Aditya Arora, Salman Khan, Munawar Hayat, Fahad~Shahbaz Khan, and Ming-Hsuan Yang.
\newblock Restormer: Efficient transformer for high-resolution image restoration.
\newblock In \emph{CVPR}, pages 5728--5739, 2022.

\bibitem[Zeng and Xie(2021)]{LLxRGTV}
Haijin Zeng and Xiaozhen Xie.
\newblock Hyperspectral image denoising via global spatial-spectral total variation regularized nonconvex local low-rank tensor approximation.
\newblock \emph{Signal Processing}, 178:\penalty0 107805, 2021.

\bibitem[Zeng et~al.(2020)Zeng, Xie, Cui, Yin, and Ning]{zeng2020hyperspectral}
Haijin Zeng, Xiaozhen Xie, Haojie Cui, Hanping Yin, and Jifeng Ning.
\newblock Hyperspectral image restoration via global l 1-2 spatial--spectral total variation regularized local low-rank tensor recovery.
\newblock \emph{IEEE TGRS}, 59\penalty0 (4):\penalty0 3309--3325, 2020.

\bibitem[Zhang et~al.(2013)Zhang, Wipf, and Zhang]{zhang2013multi}
Haichao Zhang, David Wipf, and Yanning Zhang.
\newblock Multi-image blind deblurring using a coupled adaptive sparse prior.
\newblock In \emph{CVPR}, 2013.

\bibitem[Zhang and Ghanem(2018)]{tra_3}
Jian Zhang and Bernard Ghanem.
\newblock \protect{ISTA}-net: Interpretable optimization-inspired deep network for image compressive sensing.
\newblock In \emph{CVPR}, 2018.

\bibitem[Zhang et~al.(2023)Zhang, Huang, Yao, Yang, Yu, Zhou, and Zhao]{zhang2023ingredient}
Jinghao Zhang, Jie Huang, Mingde Yao, Zizheng Yang, Hu Yu, Man Zhou, and Feng Zhao.
\newblock Ingredient-oriented multi-degradation learning for image restoration.
\newblock In \emph{CVPR}, pages 5825--5835, 2023.

\bibitem[Zhang et~al.(2017)Zhang, Zuo, Chen, Meng, and Zhang]{zhang2017beyond}
Kai Zhang, Wangmeng Zuo, Yunjin Chen, Deyu Meng, and Lei Zhang.
\newblock Beyond a gaussian denoiser: Residual learning of deep cnn for image denoising.
\newblock \emph{IEEE TIP}, 26\penalty0 (7):\penalty0 3142--3155, 2017.

\bibitem[Zhang et~al.(2018)Zhang, Zuo, and Zhang]{zhang2018ffdnet}
Kai Zhang, Wangmeng Zuo, and Lei Zhang.
\newblock Ffdnet: Toward a fast and flexible solution for cnn-based image denoising.
\newblock \emph{IEEE TIP}, 27\penalty0 (9):\penalty0 4608--4622, 2018.

\bibitem[Zhang et~al.(2024)Zhang, Ma, Wang, Zhang, Zhang, and Zhang]{zhang2024perceive}
Xu Zhang, Jiaqi Ma, Guoli Wang, Qian Zhang, Huan Zhang, and Lefei Zhang.
\newblock Perceive-ir: Learning to perceive degradation better for all-in-one image restoration.
\newblock \emph{arXiv preprint arXiv:2408.15994}, 2024.

\end{thebibliography}
}

\clearpage
\setcounter{page}{1}
\maketitlesupplementary

We present the transformer block we used in Sec. \ref{sm_sec:fc_transformer}, the framework of our hierarchical DUN in Sec. \ref{sm_sec:fc_hierarchical}, the benefits of our VLM-based degradataion module in Sec. \ref{sm_sec:fc_vlmDegra}, comparison between PGD and HQS in Sec. \ref{sm_sec:fc_pgdhqs}, computation and Params. analysis in Sec. \ref{sm_sec:compu_and_params}, comparison with other methods in Sec. \ref{sm_sec:methods}, the limitations and future work in Sec. \ref{sm_sec:limit_and_future} and more additional visual results in Sec. \ref{sm_sec:obo_more_results}.

\vspace{-2mm}
\section{Framework Clarification}
\vspace{-1mm}
\subsection{Transformer Block in VLU-Net Framework}
\label{sm_sec:fc_transformer}
\begin{figure}
    \centering
    \includegraphics[width=1\linewidth]{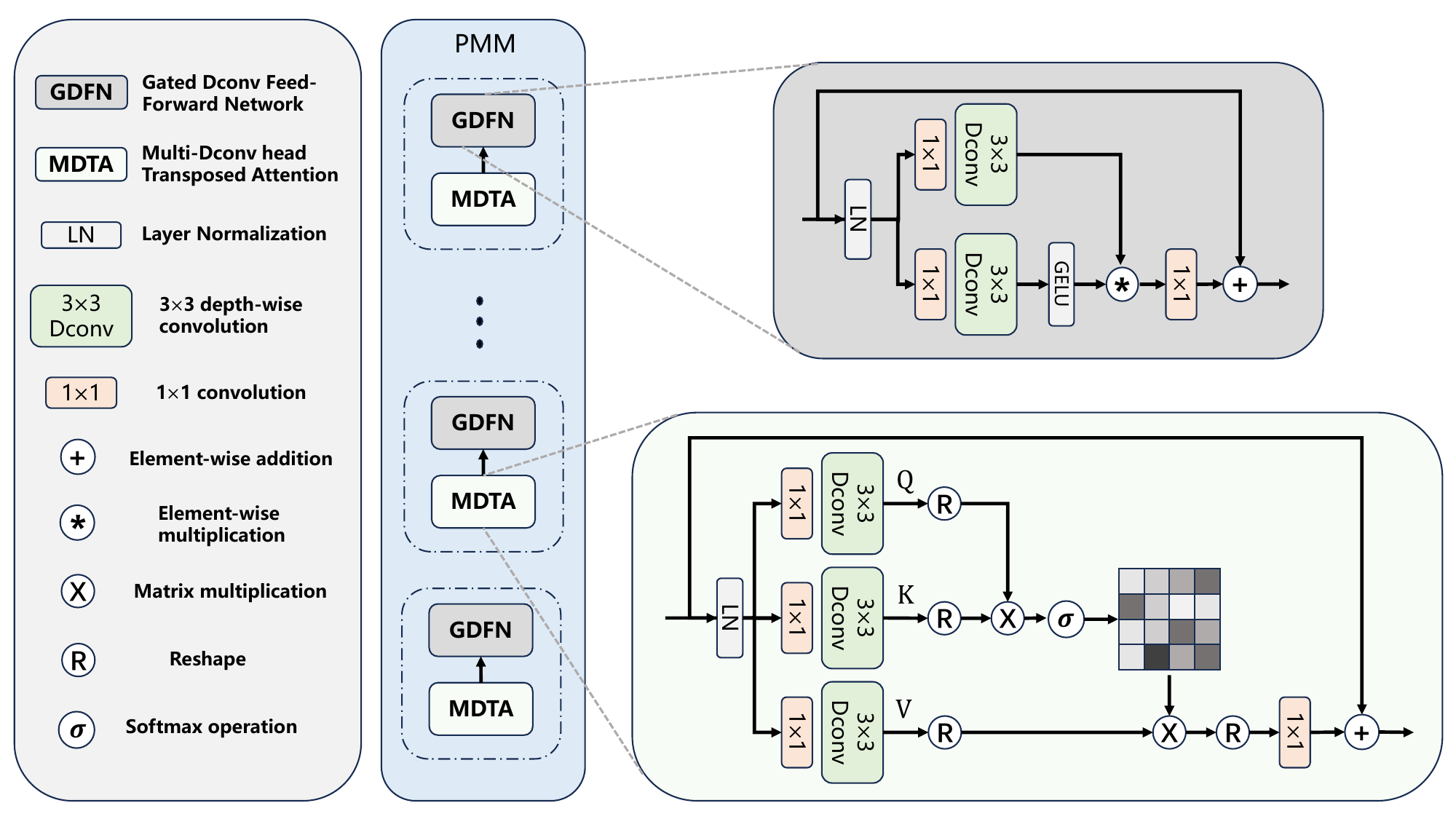}
    \vspace{-6mm}
    \caption{Overview of the Transformer block adopted in the VLU-Net framework. This Transformer block is composed of two modules, Multi-Dconv head Transposed Attention module (MDTA) and Gated-Dconv Feed-forward Network (GDFN).}
    \vspace{-6mm}
    \label{fig:transformer}
\end{figure}

\vspace{-1mm}
As mentioned in Sec. 4.1, we present Figure \ref{fig:transformer} for detailed diagram of the transformer block we adopted in the experiments following the design and hyper-parameters in [44].

Given the inputs $\hat{\mathbf{z}}^{(k)}_{(l)}$, they first passes through MDTA module, normalized by Layer Normalization (LN) and applied by a combination of one $1\times1$ convolution and one $3\times3$ depth-wise convolution.
It leads to three tensors, which are Query ($\mathbf{Q}$), Key ($\mathbf{K}$) and Value ($\mathbf{V}$).
The computation of the attention is across the channel dimensions, which effectively reduces the computational load.
Also, attention is computed in a multi-head manner in parallel.

Then the features from the MDTA are processed by the GDFN module, which consists of the LN module, the combination of one $1\times1$ convolution and one $3\times3$ depth-wise convolution.
These operations are performed through two parallel paths activated with GELU.

\vspace{-1mm}
\subsection{Framework Comparison between Current DUNs and The Proposed Hierarchical DUN}
\label{sm_sec:fc_hierarchical}
\begin{figure}
    \centering
    \includegraphics[width=1\linewidth]{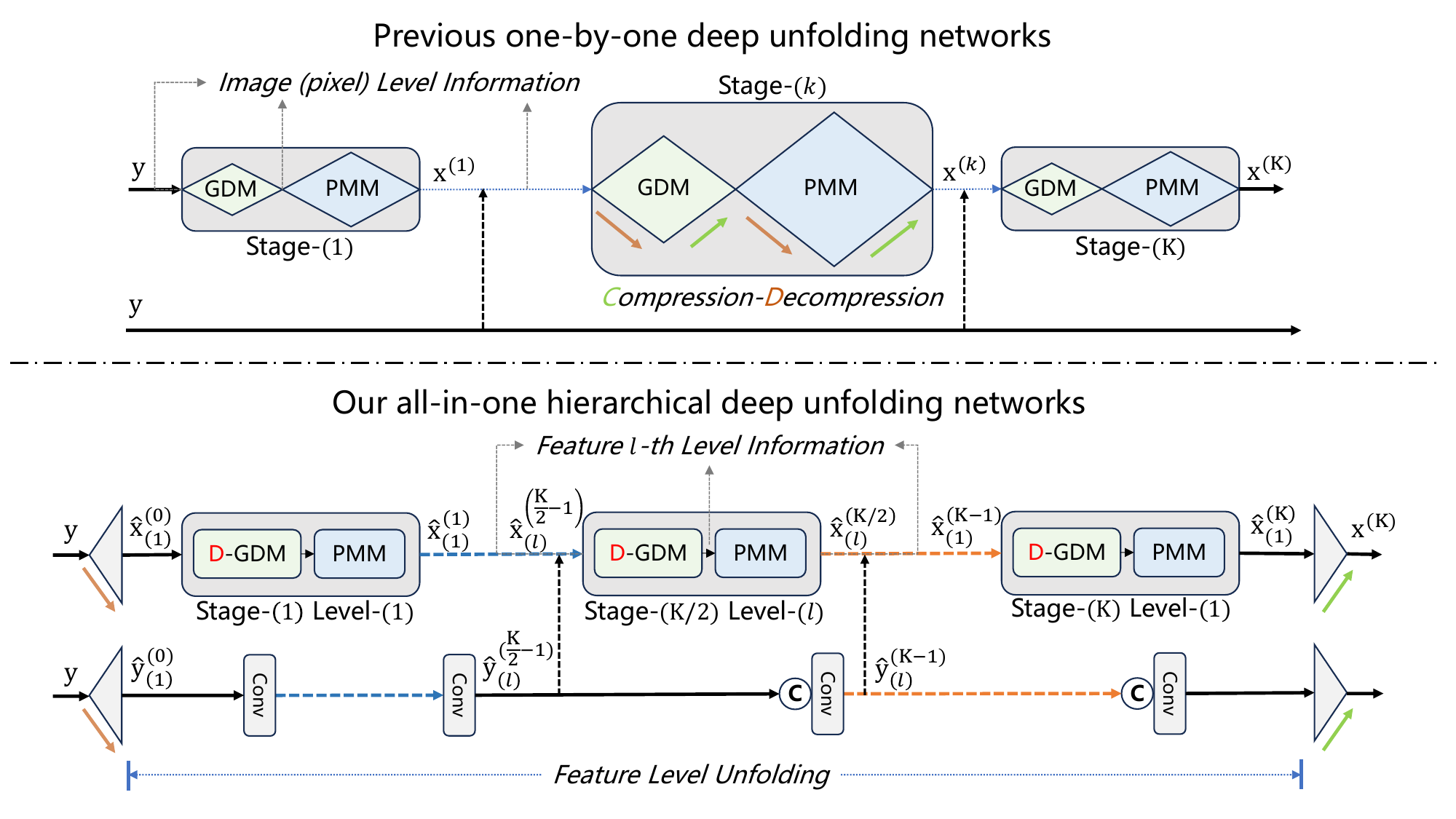}
    \vspace{-8mm}
    \caption{DUN framework comparisons. Our hierarchical DUN eliminates the loss of high-dimensional features in the compression-decompression process and fully utilizes the processing capabilities of multiple stage modules for multi-level primitive degradation and content information.}
    \vspace{-6mm}
    \label{fig:framework}
\end{figure}

\vspace{-1mm}
As mentioned in Sec. 4.3, we present Figure \ref{fig:framework} for detailed framework comparison between ours and current DUNs.

Current DUNs unfold the optimization in image level, which takes a detached view of each iteration in terms of processing objects and uses a compression-decompression strategy for information transmission between stages.
In addition, each stage actually receives the optimized image from the previous stage and the same original degraded image, which limits the ability of different stages to process from multiple perspectives and multiple levels.

Our hierarchical DUN unfolds the optimization in feature level by two linear convolution transform.
Without compromising interpretability, VLU-Net maintains high-dimensional information coherence between stages within feature unfolding strategy.
Different stages process various levels of information through sampling and convolution operations, which unleashes the potential for information processing, especially for the degradation information in GDM.

\vspace{-6mm}
\subsection{Benefits for the framework of the VLM-based Degradation Module}
\label{sm_sec:fc_vlmDegra}
\vspace{-1mm}
VLM pretrained with large image-text pairs can extract deep and general image patterns for feature enhancement.
While others such as PromptIR and Transweather, focus on shallow pixel distribution or learn queries solely from the own features.
We adopt fine-tuned VLM for degradation-text-aligned visual prompts, which automatically guide the restoration direction without additional instructions from human.
While InstructIR, using extra texts (human text instructions for inference and GPT-4 for training), learns from a semantic view and overlooks the visual degradation.

\vspace{-1mm}
\subsection{Comparison between PGD and HQS}
\label{sm_sec:fc_pgdhqs}
\vspace{-1mm}
We compare the PGD and HQS, which are the optimization algorithms for DUN framework:
PGD: gradient descent with constraint projection; HQS: auxiliary variables with alternating optimization.
Also in \cite{meng2023deep}, PGD is computationally more efficient and converges faster than HQS (e.g., the inverse operation).

\vspace{-1mm}
\section{Analysis and Comparison on the Computation and Params. of our VLU-Net}
\label{sm_sec:compu_and_params}
\vspace{-1mm}
First we provide the inference time of our VLU-Net and PromptIR for \textit{Urban100}, which is 2.7s vs. 2s per image.
Table \ref{tab:macs_comparison} shows the MACs of different methods.
\begin{table}
    \centering
\setlength{\tabcolsep}{3.0pt}
    \caption{Comparison of MACs for $256 \times 256$ input dimensions.}
    \vspace{-3mm}
\scalebox{0.85}{
    \begin{tabular}{lcccc>{\columncolor{lightpeach!50}}c}
    \toprule
        \text{Method} & InstructIR & PromptIR & MIRNet & MPRNet & \multicolumn{1}{c}{Ours}\\
    \toprule
        \text{MACs (G)} &100&160&786&588&186\\
    \bottomrule
    \end{tabular}}
    \vspace{-2mm}
    \label{tab:macs_comparison}
\end{table}
Then we analyze our hierarchical DUN structure below:
\textbf{A.} Hierarchical structure offers savings spatially due to sampling operations, especially for Transformer with square complexity.
\textbf{B.} Table \ref{tab:para_mac_block} shows our hierarchical DUN is more efficient than previous serial DUNs (expands deep features by up to 8 times (384 vs. 48), while the computational load is less than doubled (15.8 vs. 9.4)).

\begin{table}
    \centering
\setlength{\tabcolsep}{3pt}
    \caption{Params.(M, left) and MACs (G, right) of $l$-level block.}
    \vspace{-3mm}
\scalebox{0.78}{
    \begin{tabular}{lcc>{\columncolor{lightpeach!50}}cc>{\columncolor{lightpeach!50}}cc>{\columncolor{lightpeach!50}}cc>{\columncolor{lightpeach!50}}c}
    \toprule
        \text{Method} & \text{Dim} & \multicolumn{2}{c}{$l=1$} & \multicolumn{2}{c}{$l=2$} & \multicolumn{2}{c}{$l=3$} & \multicolumn{2}{c}{$l=4$}\\
    \toprule
         Serial (Previous) &48&0.2&9.4&0.2&9.4&0.2&9.4&0.2&9.4\\
         Serial (Previous)&96&0.5&\textcolor{myred}{\textbf{35.4}}&0.5&\textcolor{myred}{\textbf{35.4}}&0.5&\textcolor{myred}{\textbf{35.4}}&0.5&\textcolor{myred}{\textbf{35.4}}\\
         \rowcolor{lightgray!20}
         Hierarchical (Ours)&48-384&0.2&9.4&0.8&12.7&3.0&12.3&15.5&15.8\\
    \bottomrule
    \end{tabular}}
    \vspace{-2mm}
    \label{tab:para_mac_block}
\end{table}

\vspace{-2mm}
\section{Comparison with Competitive Methods}
\label{sm_sec:methods}
\vspace{-1mm}
\subsection{Comparison with InstructIR}
\vspace{-1mm}
\textbf{Motivation:} Ours focuses on the first all-in-one hierarchical DUN with CLIP-guided priors, while InstructIR is for VLMs' low-level abilities.
\textbf{Prompt Dependency:} InstructIR's results on low-light and blur tasks may stem from additional text prompts (real human instructions for inference and GPT-4 for training) to describe degradations, while ours relies on automatic degradation generation using only degraded inputs.
\textbf{Performance:} As evidenced in Table \ref{tab:denoising_urban100_5}, our model outperforms InstructIR among all noise levels, where the variations are challenging for humans to perceive.

\begin{table}
    \centering
\setlength{\tabcolsep}{3pt}
    \caption{Comparison for denosing with \textit{Urban100} datasets.}
    \vspace{-3mm}
\scalebox{0.8}{
    \begin{tabular}{l>{\columncolor{lightpeach!50}}cc>{\columncolor{lightpeach!50}}cc>{\columncolor{lightpeach!50}}cc>{\columncolor{lightpeach!50}}cc}
    \toprule
        \text{Method} & \multicolumn{2}{c}{$\sigma=15$} & \multicolumn{2}{c}{$\sigma=25$} & \multicolumn{2}{c}{$\sigma=50$} & \multicolumn{2}{c}{\text{Average}}\\
    \toprule
        InstructIR &33.74& .942& 31.35&.913&28.10&.852&31.06&.902\\
        \rowcolor{lightgray!20}
        VLU-Net (Ours) &\textcolor{myred}{34.25}&\textcolor{myred}{.946}& \textcolor{myred}{31.91}&\textcolor{myred}{.920}&\textcolor{myred}{28.65}&\textcolor{myred}{.864}&\textcolor{myred}{31.60}&\textcolor{myred}{.910}\\
    \bottomrule
    \end{tabular}}
    \label{tab:denoising_urban100_5}
    \vspace{-4mm}
\end{table}

\vspace{-1mm}
\subsection{Comparison with DA-CLIP}
\vspace{-1mm}
\textbf{Motivation:} Ours aims to improve hierarchical DUN, promoting degradation transform $\Phi$ for all-in-one IR.
While DA-CLIP focuses on VLMs, introducing multi-task low-level recovery abilities.
\textbf{Interpretability:} Ours originates from \textbf{optimization algorithms}, extending image to feature unfolding.
\textbf{Implementation:} We differ in fine-tuning strategies (lightweight MLP adapters vs. ControlNet) and basic models (DUN with Restormer vs. Diffusion).

\vspace{-1mm}
\subsection{Comparison with DGUNet}
\vspace{-1mm}
\textbf{Restoration Objective:} Ours aims to use a unified model to recover low-quality images in a multi-task all-in-one situation, whereas DGUNet can only function under a single task (one-by-one) and needs to train separate models under different datasets to be applied to multiple recovery tasks.
\textbf{All-in-one Strategy:} Ours uses the VLM to provide visual prior and guide the construction of degradation transforms, discriminate and adapt multiple degradation inputs, and support all-in-one IR.
Whereas, DGUNet utilizes learnable convolutions to construct a single degradation transform, which can only learn one single degradation pattern a training time.
\textbf{DUN Structure:} We use a hierarchical DUN approach to build the model from image unfolding to feature unfolding, while DGUNet concatenates each iteration block at the image level.
By comparing Table \ref{tab:ablation_NHRBL} and Table \ref{tab:para_mac_block}, we can see that our hierarchical DUN has advantages in the number of parameters, computation and results.

\vspace{-2mm}
\section{Limitations and Future Work}
\label{sm_sec:limit_and_future}
\vspace{-1mm}
Although our VLU-Net efficiently models different degradations by leveraging VLM into hierarchical DUN framework with superior performance in five degradation NHRBL and three degradation NHR, it is unclear how its performance with other corruptions such as snow or JPEG artifacts.
In addition, it is also worthy to explore more advanced potential of GDM in DUN, especially in the proposed feature-level hierarchical DUN architecture.
Classification is the first step of recovery to provide guidance, while the recovery results depend on estimating degradation transform $\Phi$ and solving the ill-posed inverse problem, which are also challenging. 
For the finetuning strategy (single positive sample in the numerator) we used (Eq.~\ref{loss_single_sample}) in the model, which may result in inconsistencies when degradations are rare.
Attempts such as controling the ratio of degraded samples and using multiple positive sample loss may be adopted for our future work.

\vspace{-2mm}
\section{Addition Visual Results}
\vspace{-1mm}
\subsection{Visual Degradation of Blur}
\begin{figure}
    \centering
    \includegraphics[width=0.32\linewidth]{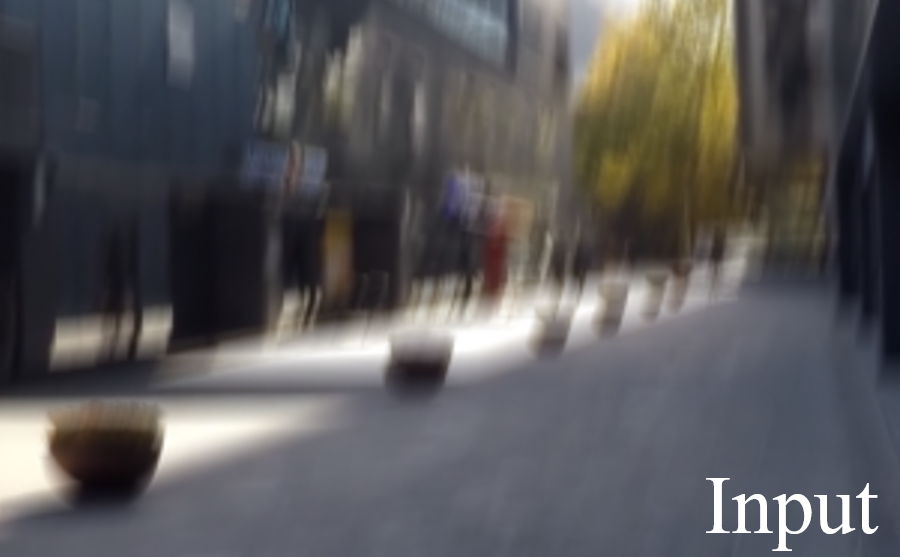}
    \includegraphics[width=0.32\linewidth]{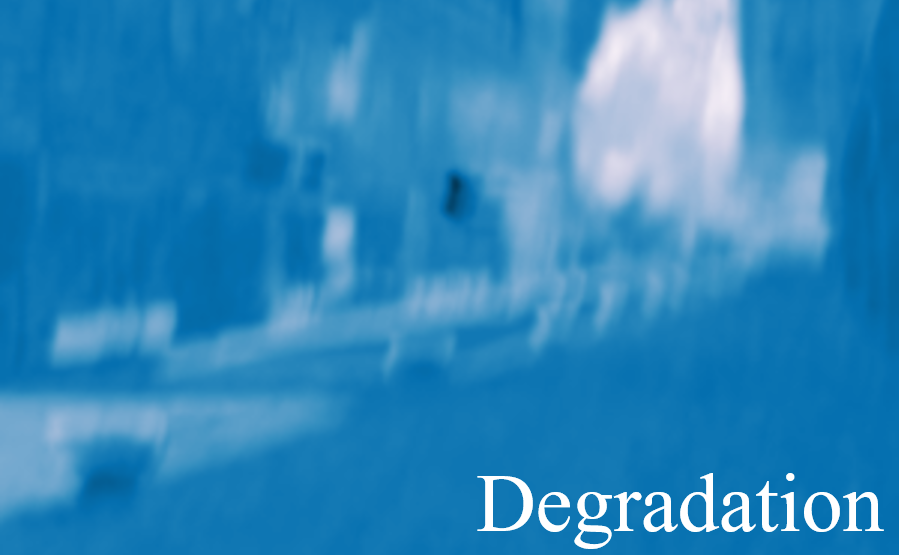}
    \includegraphics[width=0.32\linewidth]{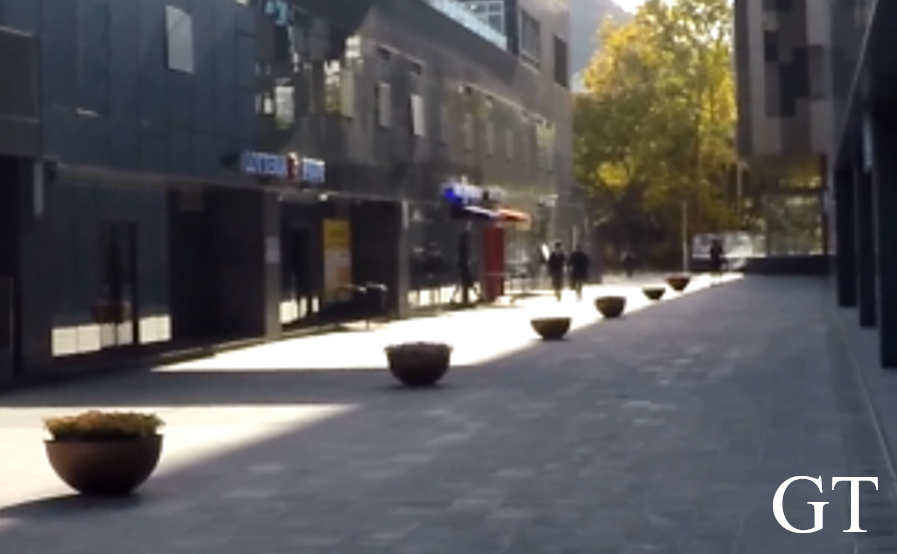}\\
    \vspace{-2mm}
    \caption{Visual degradation from D-GDM for blur.}
    \label{fig:visual_degradation_blur}
    \vspace{-6mm}
\end{figure}

\vspace{-1mm}
We included Figure~\ref{fig:visual_degradation_blur} to illustrate the blur degradation. 
Here, we adopt attention modules for the degradation transform $\Phi$ to help better approximate and learn the convolution pattern.

\vspace{-1mm}
\subsection{More Qualitative Results for One-by-one IR}
\label{sm_sec:obo_more_results}
\begin{figure*}
    \centering
    \setlength{\tabcolsep}{0mm}
    \renewcommand{\arraystretch}{0}
    \begin{tabular}{lccccc}
        \rotatebox{90}{    \quad Degraded}&
        \includegraphics[width=0.18\linewidth]{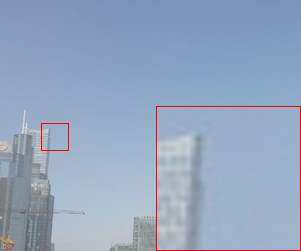}&
        \includegraphics[width=0.18\linewidth]{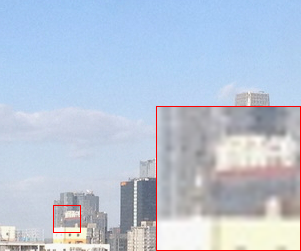}&
        \includegraphics[width=0.18\linewidth]{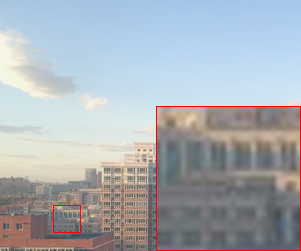}&
        \includegraphics[width=0.18\linewidth]{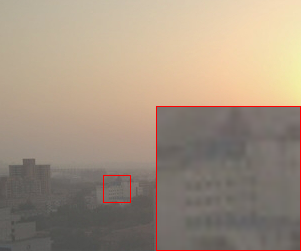}&
        \includegraphics[width=0.18\linewidth]{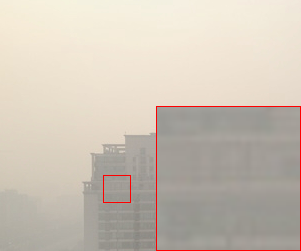}\\
        \rotatebox{90}{    \quad InstructIR}&
        \includegraphics[width=0.18\linewidth]{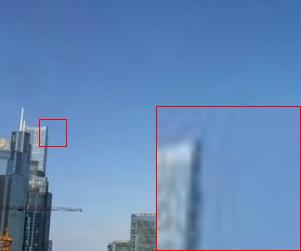}&
        \includegraphics[width=0.18\linewidth]{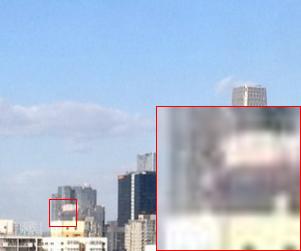}&
        \includegraphics[width=0.18\linewidth]{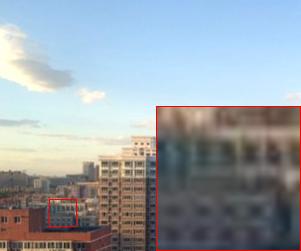}&
        \includegraphics[width=0.18\linewidth]{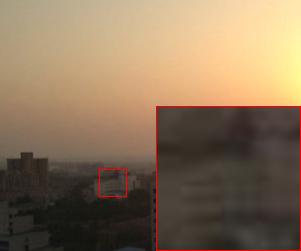}&
        \includegraphics[width=0.18\linewidth]{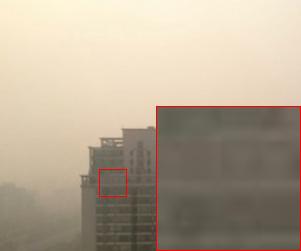}\\
        \rotatebox{90}{     \textbf{VLU-Net(Ours)}}&
        \includegraphics[width=0.18\linewidth]{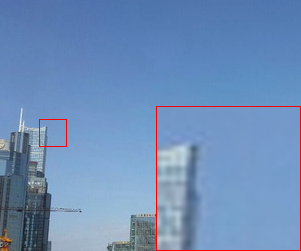}&
        \includegraphics[width=0.18\linewidth]{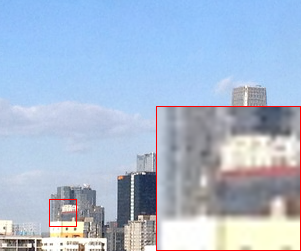}&
        \includegraphics[width=0.18\linewidth]{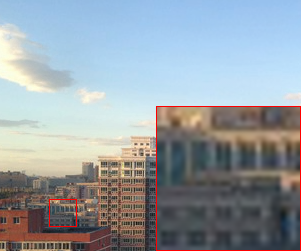}&
        \includegraphics[width=0.18\linewidth]{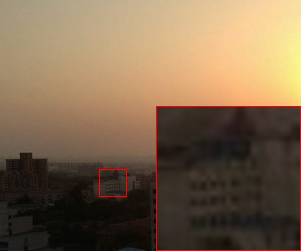}&
        \includegraphics[width=0.18\linewidth]{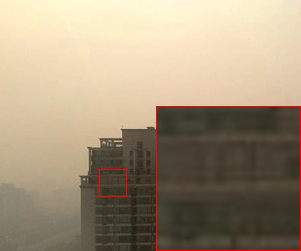}\\
        \centering \rotatebox{90}{\quad \quad \quad GT}&
        \includegraphics[width=0.18\linewidth]{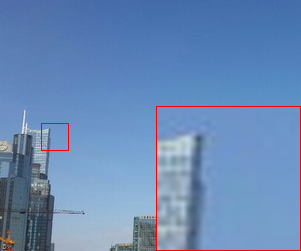}&
        \includegraphics[width=0.18\linewidth]{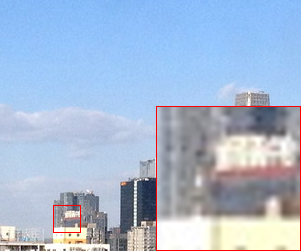}&
        \includegraphics[width=0.18\linewidth]{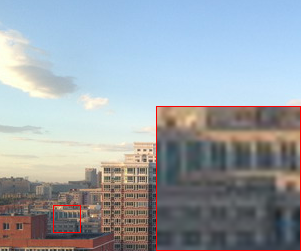}&
        \includegraphics[width=0.18\linewidth]{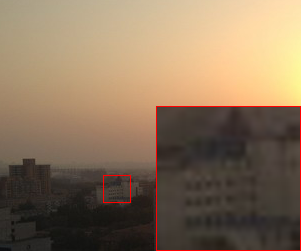}&
        \includegraphics[width=0.18\linewidth]{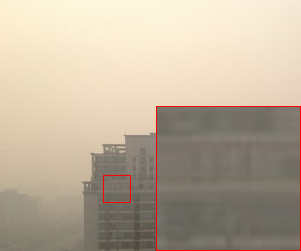}\\
    \end{tabular}
    \vspace{-2mm}
    \caption{Dehazing results of one-by-one IR.}
    \vspace{-1mm}
    \label{fig:single_dehazing}
\end{figure*}

\begin{figure*}
    \centering
    \setlength{\tabcolsep}{0mm}
    \renewcommand{\arraystretch}{0}
    \begin{tabular}{lccccc}
        \rotatebox{90}{    \quad Degraded}&
        \includegraphics[width=0.18\linewidth]{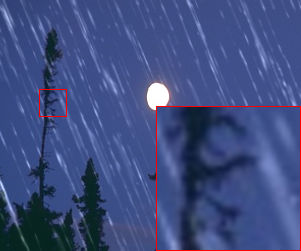}&
        \includegraphics[width=0.18\linewidth]{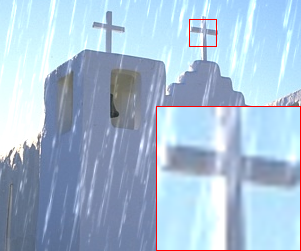}&
        \includegraphics[width=0.18\linewidth]{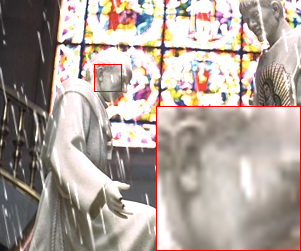}&
        \includegraphics[width=0.18\linewidth]{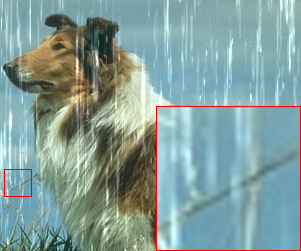}&
        \includegraphics[width=0.18\linewidth]{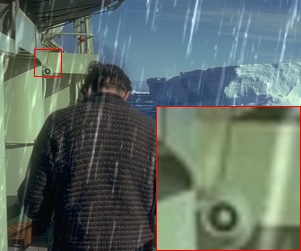}\\
        \rotatebox{90}{    \quad InstructIR}&
        \includegraphics[width=0.18\linewidth]{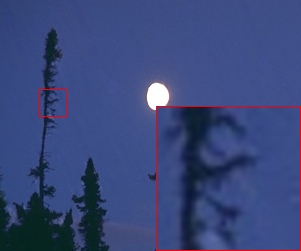}&
        \includegraphics[width=0.18\linewidth]{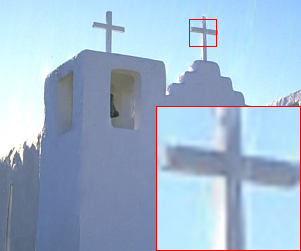}&
        \includegraphics[width=0.18\linewidth]{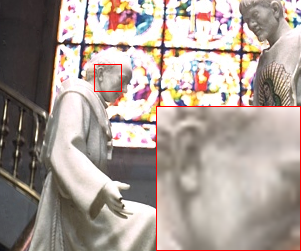}&
        \includegraphics[width=0.18\linewidth]{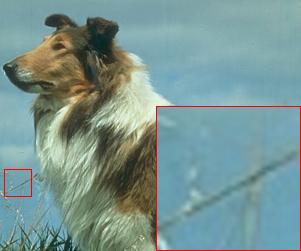}&
        \includegraphics[width=0.18\linewidth]{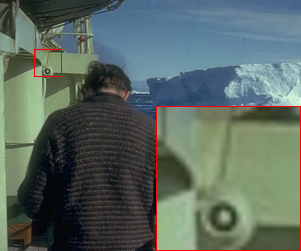}\\
        \rotatebox{90}{     \textbf{VLU-Net(Ours)}}&
        \includegraphics[width=0.18\linewidth]{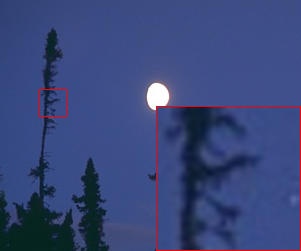}&
        \includegraphics[width=0.18\linewidth]{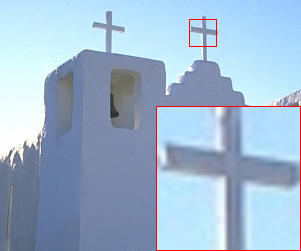}&
        \includegraphics[width=0.18\linewidth]{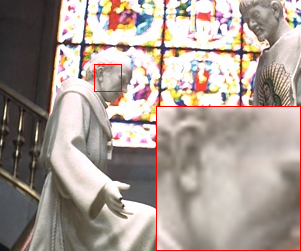}&
        \includegraphics[width=0.18\linewidth]{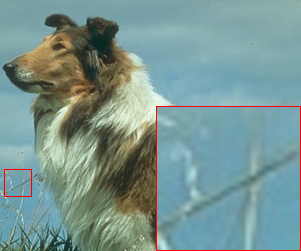}&
        \includegraphics[width=0.18\linewidth]{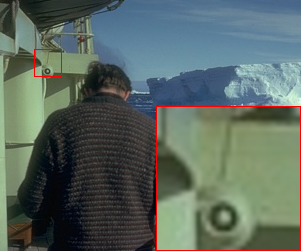}\\
        \centering \rotatebox{90}{\quad \quad \quad GT}&
        \includegraphics[width=0.18\linewidth]{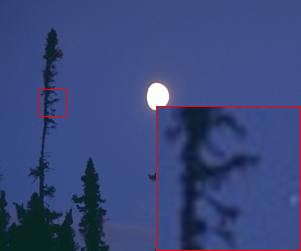}&
        \includegraphics[width=0.18\linewidth]{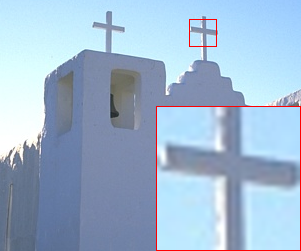}&
        \includegraphics[width=0.18\linewidth]{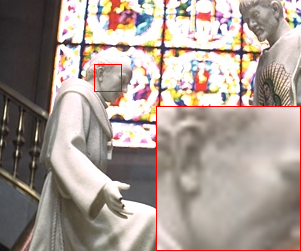}&
        \includegraphics[width=0.18\linewidth]{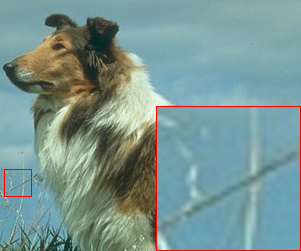}&
        \includegraphics[width=0.18\linewidth]{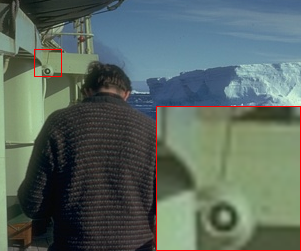}\\
    \end{tabular}
    \vspace{-2mm}
    \caption{Deraining results of one-by-one IR.}
    \label{fig:single_deraining}
\end{figure*}

\begin{figure*}
    \centering
    \setlength{\tabcolsep}{0mm}
    \renewcommand{\arraystretch}{0}
    \begin{tabular}{lccccc}
        \rotatebox{90}{    \quad Degraded}&
        \includegraphics[width=0.18\linewidth]{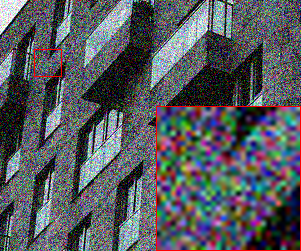}&
        \includegraphics[width=0.18\linewidth]{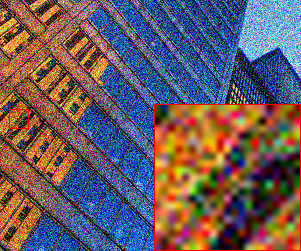}&
        \includegraphics[width=0.18\linewidth]{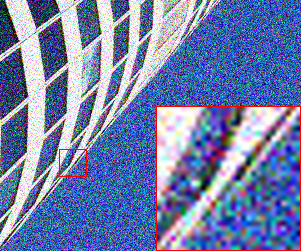}&
        \includegraphics[width=0.18\linewidth]{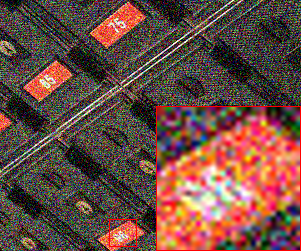}&
        \includegraphics[width=0.18\linewidth]{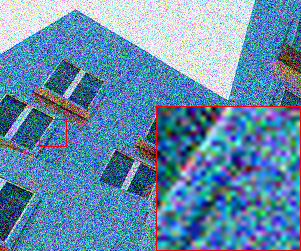}\\
        \rotatebox{90}{    \quad InstructIR}&
        \includegraphics[width=0.18\linewidth]{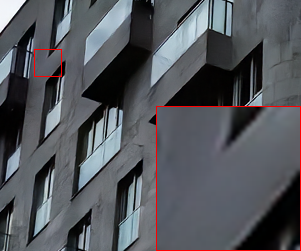}&
        \includegraphics[width=0.18\linewidth]{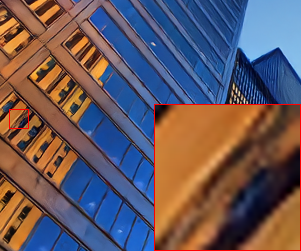}&
        \includegraphics[width=0.18\linewidth]{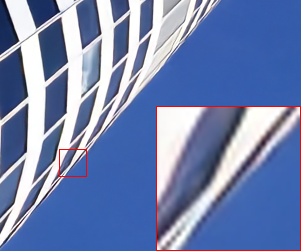}&
        \includegraphics[width=0.18\linewidth]{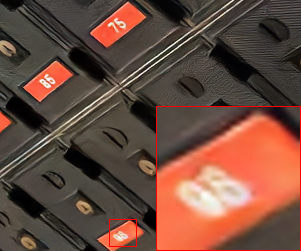}&
        \includegraphics[width=0.18\linewidth]{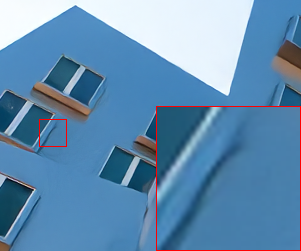}\\
        \rotatebox{90}{     \textbf{VLU-Net(Ours)}}&
        \includegraphics[width=0.18\linewidth]{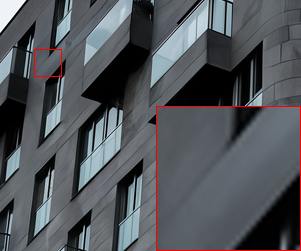}&
        \includegraphics[width=0.18\linewidth]{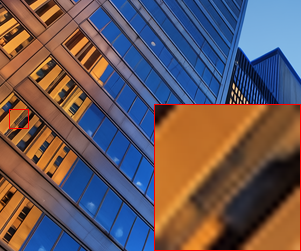}&
        \includegraphics[width=0.18\linewidth]{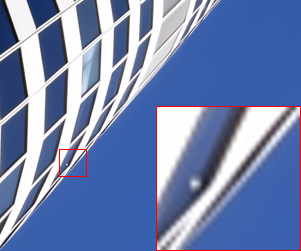}&
        \includegraphics[width=0.18\linewidth]{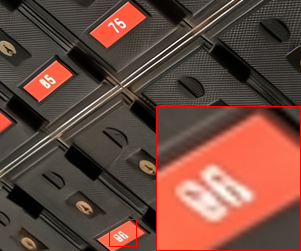}&
        \includegraphics[width=0.18\linewidth]{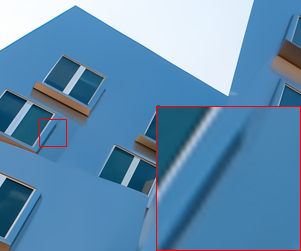}\\
        \centering \rotatebox{90}{\quad \quad \quad GT}&
        \includegraphics[width=0.18\linewidth]{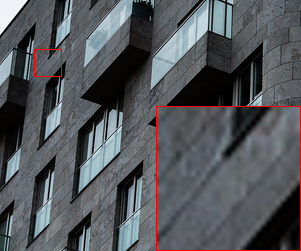}&
        \includegraphics[width=0.18\linewidth]{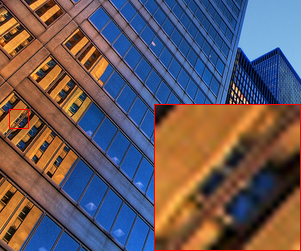}&
        \includegraphics[width=0.18\linewidth]{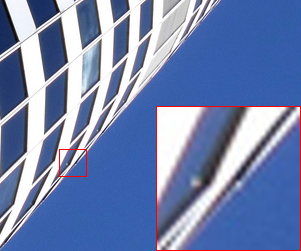}&
        \includegraphics[width=0.18\linewidth]{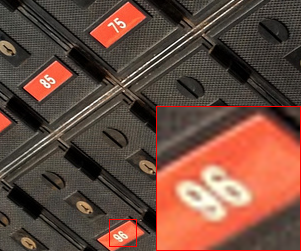}&
        \includegraphics[width=0.18\linewidth]{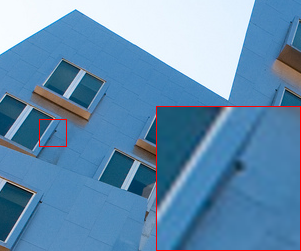}\\
    \end{tabular}
    \caption{Denoising results of one-by-one IR.}
    \label{fig:single_denoising}
\end{figure*}

\vspace{-1mm}
We present more qualitative results of one-by-one IR (Dehazing in Figure \ref{fig:single_dehazing}, Deraining in Figure \ref{fig:single_deraining} and Denoising with $\sigma = 50$ in Figure \ref{fig:single_denoising}, which further elucidate the effectiveness of our VLU-Net under the single-task setting.
We crop and zoom in the images for better presentation.

\end{document}